\documentclass{article}
\usepackage[left=3cm,right=3cm,bottom=3cm,top=3cm]{geometry}

\usepackage{microtype}
\usepackage{graphicx}
\usepackage{booktabs} %

\usepackage{nicefrac}       %
\usepackage{xcolor}         %

\usepackage{tikz}
\usetikzlibrary{backgrounds,calc}
\usepackage{mathtools}

\usepackage{amsthm}  %
\usepackage{amsmath}
\usepackage{amsfonts}
\usepackage{thmtools}  %

\usepackage{bbm}
\usepackage[export]{adjustbox}
\usepackage{tablefootnote}
\usepackage{multirow}
\usepackage{algorithm}%
\usepackage{algpseudocode}%
\usepackage{amsfonts}
\usepackage{makecell}
\usepackage{graphicx}
\usepackage{subcaption}
\usepackage{amsthm}
\usepackage{wrapfig}
\DeclareMathOperator*{\argmax}{arg\,max}

\DeclarePairedDelimiterX{\norm}[1]{\lVert}{\rVert}{#1}

\usepackage[most]{tcolorbox}
\usepackage{efbox}

\usepackage[pagebackref=true]{hyperref}
\usepackage{cleveref}

\usepackage[numbers]{natbib}
\title{Metadata Archaeology: Unearthing Data Subsets \\ by Leveraging Training Dynamics \vspace{1cm}}

\author{
  Shoaib Ahmed Siddiqui \\
  University of Cambridge\\
  \texttt{msas3@cam.ac.uk} \\
  \and
  Nitarshan Rajkumar \\
  University of Cambridge\\
  \texttt{nr500@cam.ac.uk} \\
  \and
  Tegan Maharaj \\
  University of Toronto\\
  \texttt{tegan.maharaj@utoronto.ca} \\
  \and
  David Krueger \\
  University of Cambridge\\
  \texttt{dsk30@cam.ac.uk} \\
  \and
  Sara Hooker \\
  Cohere for AI\\
  \texttt{sarahooker@cohere.com} \\
}

\date{}

\begin{document}

\maketitle

\begin{abstract}

Modern machine learning research relies on relatively few carefully curated datasets. Even in these datasets, and typically in `untidy' or raw data, practitioners are faced with significant issues of data quality and diversity which can be prohibitively labor intensive to address. 
Existing methods for dealing with these challenges tend to make strong assumptions about the particular issues at play, and often require \textit{a priori} knowledge or metadata such as domain labels. Our work is orthogonal to these methods: we instead focus on providing a unified and efficient framework for \textit{Metadata Archaeology} -- uncovering and inferring metadata of examples in a dataset.
We curate different subsets of data that might exist in a dataset (e.g.\ mislabeled, atypical, or out-of-distribution examples) using simple transformations, and leverage differences in learning dynamics between these \textbf{probe suites} to infer metadata of interest. 
Our method is on par with far more sophisticated mitigation methods across different tasks: identifying and correcting mislabeled examples, classifying minority-group samples, prioritizing points relevant for training and enabling scalable human auditing of relevant examples. 
\end{abstract}

\section{Introduction}\label{sec:intro}
Modern machine learning is characterized by ever-larger datasets and models. The expanding scale has produced impressive progress \cite{Wei2022,Kaplan2020,roberts2020} yet presents both optimization and auditing challenges. Real-world dataset collection techniques often result in significant label noise \citep{2022vasudevan}, and can present significant numbers of redundant, corrupted, or duplicate inputs \citep{carlini2022}. Scaling the size of our datasets makes detailed human analysis and auditing labor-intensive, and often simply infeasible. These realities motivate a consideration of how to efficiently characterize different aspects of the data distribution.

Prior work has developed a rough taxonomy of data properties, or \textbf{metadata} which different examples might exhibit, including but not limited to: \textit{noisy} \citep{wu2020topological,yi2019probabilistic,pmlr-v97-thulasidasan19a,Thulasidasan2019CombatingLN}, \textit{atypical} \citep{hooker2020characterising,pmlr-v81-buolamwini18a,hashimoto18a,Sowik2021AlgorithmicBA}, \textit{challenging} \citep{hooker2019,ahia-etal-2021-low-resource,Baldock2021DeepLT,paul2021deep,agarwal2021estimating}, \textit{prototypical or core subset selection} \citep{paul2021deep,sener2018active,Shim2021CoresetSF,huggins2017coresets} and \textit{out-of-distribution} \citep{LeBrun2022EvaluatingDD}. While important progress has been made on some of these metadata categories individually, these categories are typically addressed in isolation reflecting an overly strong assumption that only one, known issue is at play in a given dataset. %

For example, considerable work has focused on the issue of label noise. A simple yet widely-used approach to mitigate label noise is to remove the impacted data examples \citep{pleiss2020identifying}. However, it has been shown that it is challenging to distinguish difficult examples from noisy ones, which often leads to useful data being thrown away when both noisy and atypical examples are present \citep{Wang2018IterativeLW,talukdar-etal-2021-training}. 

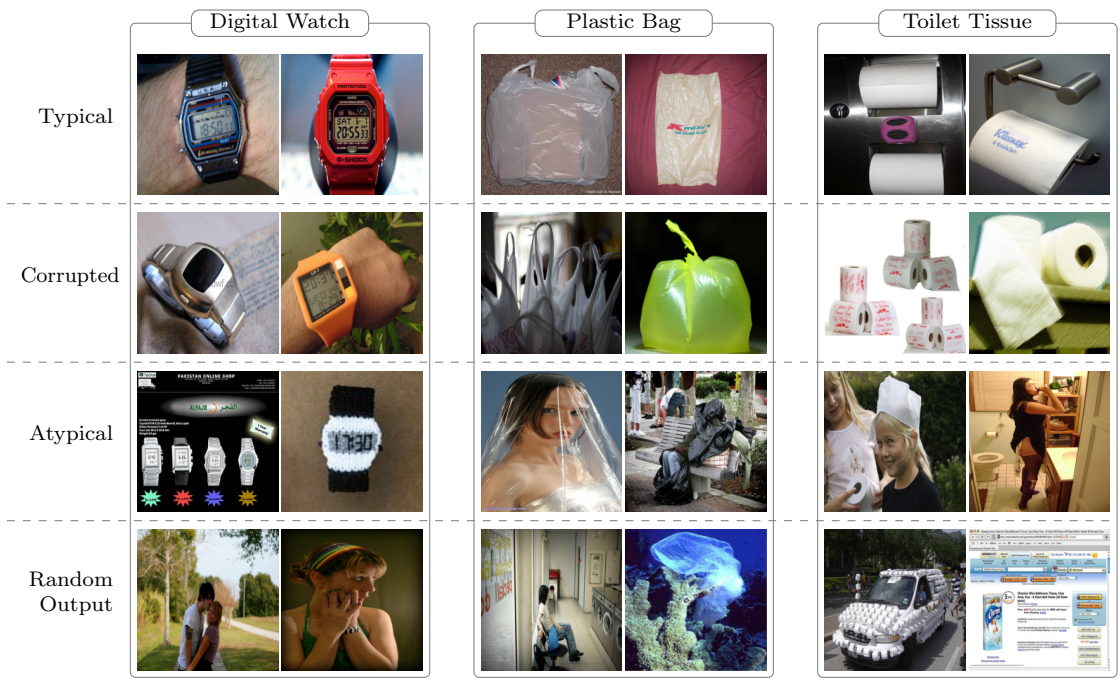
\begin{figure*}[!t]\centering
    \begin{tikzpicture}[x=130pt,y=-60pt,
            head/.style={font=\footnotesize,execute at begin node={\strut},execute at end node={\strut}},
            cn/.style={head,text width=65pt,align=center,draw,rectangle,rounded corners=3pt,inner xsep=1pt,inner ysep=0.25pt,fill=white,draw=black!50},
            rn/.style={head,text width=35pt,inner sep=5pt,align=right},
            img/.style={inner sep=0.5pt,yshift=25pt},
            L/.style={img,anchor=north east},
            R/.style={img,anchor=north west},
        ]
        \def\colimg#1{\includegraphics[width=0.12\linewidth]{Figures/loss_traj/surfacing/imagenet/ex_surfaced/#1}}
            \node[rn] (r1) at (0.38,1) {Typical};
            \node[rn] (r2) at (0.38,2) {Corrupted};
            \node[rn] (r3) at (0.38,3) {Atypical};
            \node[rn] (r4) at (0.38,4) {Random\\ Output};
            \node[cn] (c1) at (1,0.4) {Digital Watch};
            \node[cn] (c2) at (2,0.4) {Plastic Bag};
            \node[cn] (c3) at (3,0.4) {Toilet Tissue};
            \node[L] (e-1-1) at (r1-|c1.south) {\colimg{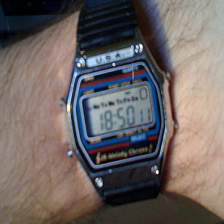}};
            \node[R] (e-2-1) at (r1-|c1.south) {\colimg{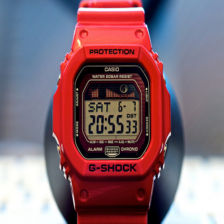}};
            \node[L] (e-3-1) at (r1-|c2.south) {\colimg{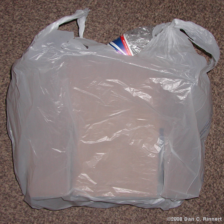}};
            \node[R] (e-4-1) at (r1-|c2.south) {\colimg{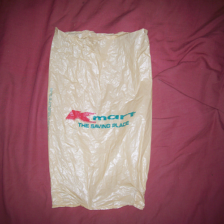}};
            \node[L] (e-5-1) at (r1-|c3.south) {\colimg{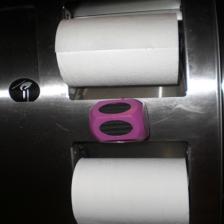}};
            \node[R] (e-6-1) at (r1-|c3.south) {\colimg{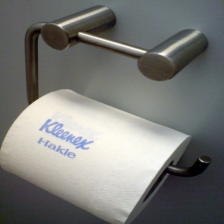}};
            \node[L] (e-1-2) at (r2-|c1.south) {\colimg{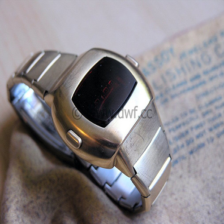}};
            \node[R] (e-2-2) at (r2-|c1.south) {\colimg{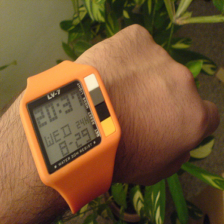}};
            \node[L] (e-3-2) at (r2-|c2.south) {\colimg{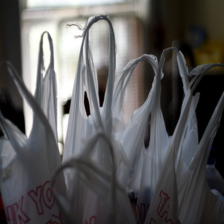}};
            \node[R] (e-4-2) at (r2-|c2.south) {\colimg{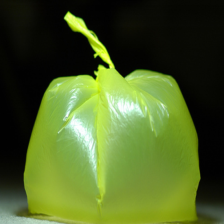}};
            \node[L] (e-5-2) at (r2-|c3.south) {\colimg{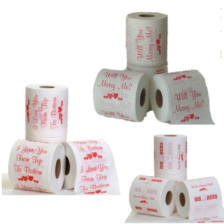}};
            \node[R] (e-6-2) at (r2-|c3.south) {\colimg{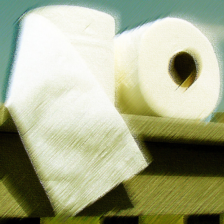}};
            \node[L] (e-1-3) at (r3-|c1.south) {\colimg{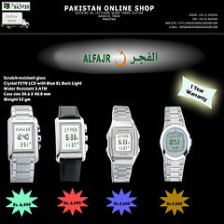}};
            \node[R] (e-2-3) at (r3-|c1.south) {\colimg{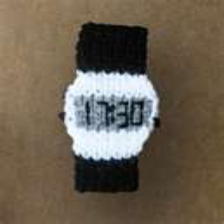}};
            \node[L] (e-3-3) at (r3-|c2.south) {\colimg{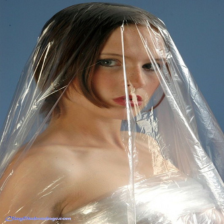}};
            \node[R] (e-4-3) at (r3-|c2.south) {\colimg{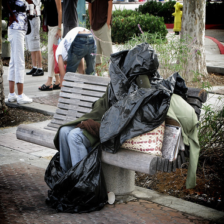}};
            \node[L] (e-5-3) at (r3-|c3.south) {\colimg{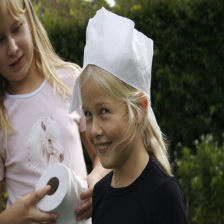}};
            \node[R] (e-6-3) at (r3-|c3.south) {\colimg{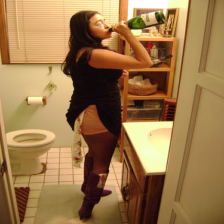}};
            \node[L] (e-1-4) at (r4-|c1.south) {\colimg{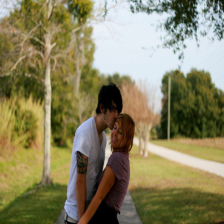}};
            \node[R] (e-2-4) at (r4-|c1.south) {\colimg{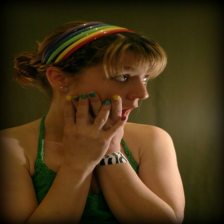}};
            \node[L] (e-3-4) at (r4-|c2.south) {\colimg{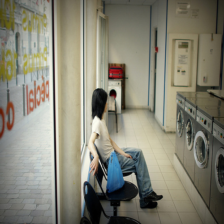}};
            \node[R] (e-4-4) at (r4-|c2.south) {\colimg{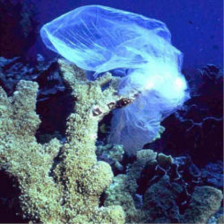}};
            \node[L] (e-5-4) at (r4-|c3.south) {\colimg{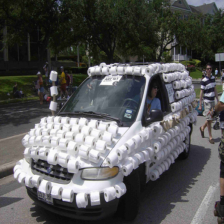}};
            \node[R] (e-6-4) at (r4-|c3.south) {\colimg{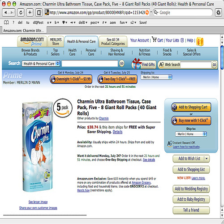}};
            \begin{scope}[on background layer,rounded corners=2pt]
                \draw[black!50] ([shift={(-2pt,0pt)}]c1.west-|e-1-1.west) rectangle ([shift={(2pt,-2pt)}]e-2-4.south east);
                \draw[black!50] ([shift={(-2pt,0pt)}]c2.west-|e-3-1.west) rectangle ([shift={(2pt,-2pt)}]e-4-4.south east);
                \draw[black!50] ([shift={(-2pt,0pt)}]c3.west-|e-5-1.west) rectangle ([shift={(2pt,-2pt)}]e-6-4.south east);
                \coordinate (R1) at ($(e-1-1.south)!0.5!(e-1-2.north)$);
                \coordinate (R2) at ($(e-1-2.south)!0.5!(e-1-3.north)$);
                \coordinate (R3) at ($(e-1-3.south)!0.5!(e-1-4.north)$);
                \draw[dashed,black!50] (R1-|r1.west)--([xshift=2pt]R1-|e-6-1.east);
                \draw[dashed,black!50] (R2-|r1.west)--([xshift=2pt]R2-|e-6-1.east);
                \draw[dashed,black!50] (R3-|r1.west)--([xshift=2pt]R3-|e-6-1.east);
            \end{scope}
    \end{tikzpicture}
    \caption{Examples surfaced through the use of \textit{MAP-D} on ImageNet train set. \textbf{Column} title is the ground truth class, \textbf{row} title is the  metadata category assigned by \textit{MAP-D}.\textit{ MAP-D} performs metadata archaeology by curating a probe set and then probing for similar examples based on training dynamics. This approach can bring to light biases, mislabelled examples, and other dataset issues.
    \vspace{-2mm}
    }
    \label{fig:surfaced_examples_imagenet}
\end{figure*}

Meanwhile, loss-based prioritization \citep{jiang2019,Katharopoulos2018} techniques essentially do the opposite -- these techniques \textit{upweight} high loss examples, assuming these examples are challenging yet learnable. These methods have been shown to quickly degrade in performance in the presence of even small amounts of noise since upweighting noisy samples hurts generalization \citep{hu2021does,paul2021deep}.

The underlying issue with such approaches is the assumption of a single, known type of data issue. Interventions are often structured to identify examples as simple vs. challenging, clean vs. noisy, typical vs. atypical, in-distribution vs. out-of-distribution etc. However, large scale datasets may present subsets with many different properties. In these settings, understanding the interactions between an intervention and many different subsets of interest can help prevent points of failure. Moreover, relaxing the notion that all these properties are treated independently allows us to capture realistic scenarios where multiple metadata annotations can apply to the same datapoint. For example, a \textit{challenging} example may be so because it is \textit{atypical}.

In this work, we are interested in moving away from a siloed treatment of different data properties. We use the term \textbf{Metadata Archaeology} to describe the problem of inferring metadata across a more complete data taxonomy. Our approach, which we term \textbf{Metadata Archaeology via Probe Dynamics (\textit{MAP-D})}, leverages distinct differences in training dynamics for different curated subsets to enable specialized treatment and effective labelling of different metadata categories. Our methods of constructing these probes are general enough that the same probe category can be crafted efficiently for many different datasets with limited domain-specific knowledge.

We present consistent results across six image classification datasets, CIFAR-10/CIFAR-100 \citep{krizhevsky2009learning}, ImageNet \citep{imagenet_cvpr09}, Waterbirds \citep{sagawa2020investigation}, CelebA \citep{liu2015faceattributes} , Clothing1M \citep{xiao2015clothing1m} and two models from the ResNet family \citep{he2016deep}. Our simple approach is competitive with far more complex mitigation techniques designed to only treat one type of metadata in isolation. Furthermore, it outperforms other methods in settings dealing with multiple sources of uncertainty simultaneously.
We summarize our contributions as:

\begin{itemize}
    \item We propose \textbf{Metadata Archeology}, a unifying and general framework for uncovering latent metadata categories.%
    \item We introduce and validate the approach of \textbf{Metadata Archaeology via Probe Dynamics (\textit{MAP-D})}: leveraging the training dynamics of curated data subsets called \textbf{probe suites} to infer other examples' metadata.
    \item We show how \textit{MAP-D} could be leveraged to audit large-scale datasets or debug model training, with negligible added cost - see \Cref{fig:surfaced_examples_imagenet}.
    This is in contrast to prior work which presents a siloed treatment of different data properties.
    \item We use \textit{MAP-D} to identify and correct mislabeled examples in a dataset. Despite its simplicity, \textit{MAP-D} is on-par with far more sophisticated methods, while enabling natural extension to an arbitrary number of modes. 
    \item Finally, we show how to use \textit{MAP-D} to identify minority group samples, or surface examples for data-efficient prioritized training.

\end{itemize}

\section{Metadata Archaeology via Probe Dynamics (MAP-D)}\label{sec:metadata_overview}

Metadata is data about data, for instance specifying when, where, or how an example was collected. This could include the provenance of the data, or information about its quality (e.g.\ indicating that it has been corrupted by some form of noise).
An important distinguishing characteristic of metadata is that it can be \textit{relational}, explaining how an example compares to others. For instance, whether an example is typical or atypical, belongs to a minority class, or is out-of-distribution (OOD), are all dependent on the entire data distribution.

The problem of \textbf{metadata archaeology} is the inference of metadata $m \subset \mathcal{M}$ given a dataset $\mathcal{D}$. While methods for inferring $m$ might also be useful for semi-supervised labelling or more traditional feature engineering, and vice versa, it is the relational nature of metadata that makes this problem unique and often computationally expensive.

\subsection{Methodology}\label{sec:methodology}

\textbf{Metadata Archaeology via Probe Dynamics (\textit{MAP-D})}, leverages differences in the statistics of learning curves across metadata features to infer the metadata of previously unseen examples.

We consider a model which learns a function  $f_{\theta}: \mathcal{X} \mapsto \mathcal{Y}$ with trainable weights $\theta$. Given the training dataset $\mathcal{D}$, $f_{\theta}$ optimizes a set of weights $\theta^*$ by minimizing an objective function $L$ with loss $l$ for each example.

We assume that the learner has access to two types of samples for training.
\emph{First} is a training set $\mathcal{D}$:

\begin{equation}
\mathcal{D} := \big\{(x_{1}, y_{1}),\dots ,(x_{N}, y_{N})\big\} \subset \mathcal{X} \times \mathcal{Y}
\end{equation}

\noindent where $\mathcal{X}$ represents the data space and $\mathcal{Y}$ the set of outcomes associated with the respective instances. Examples in the data space are also assumed to have associated, but unobserved, metadata $m \subset \mathcal{M}$.

\emph{Secondly}, we assume the learner to also have access to a small curated subset of $j$ samples ($j \leq N$; typically $j \ll N$) associated with metadata $m \subset \mathcal{M}$, i.e.:

\begin{equation}
\mathcal{D}_{m} := \{(x_{1}, y_{1}, m_{1}),\dots ,(x_{j}, y_{j}, m_{j})\}  \subset \mathcal{X} \times \mathcal{Y} \times \mathcal{M} \enspace
\end{equation}

We refer to these curated subsets as probe suites. A key criteria is for our method to require very few annotated probe examples ($j \ll N$). In this work, we focus on probe suits which can be constructed algorithmically, as human annotations of metadata require costly human effort to maintain.

\subsubsection{Assigning Metadata Features to Unseen Examples}
\label{subsec:assigning_examples}

\begin{figure*}[!t]
    \centering
    \subfloat[Probe Accuracy]{
        \includegraphics[width=0.32\linewidth]{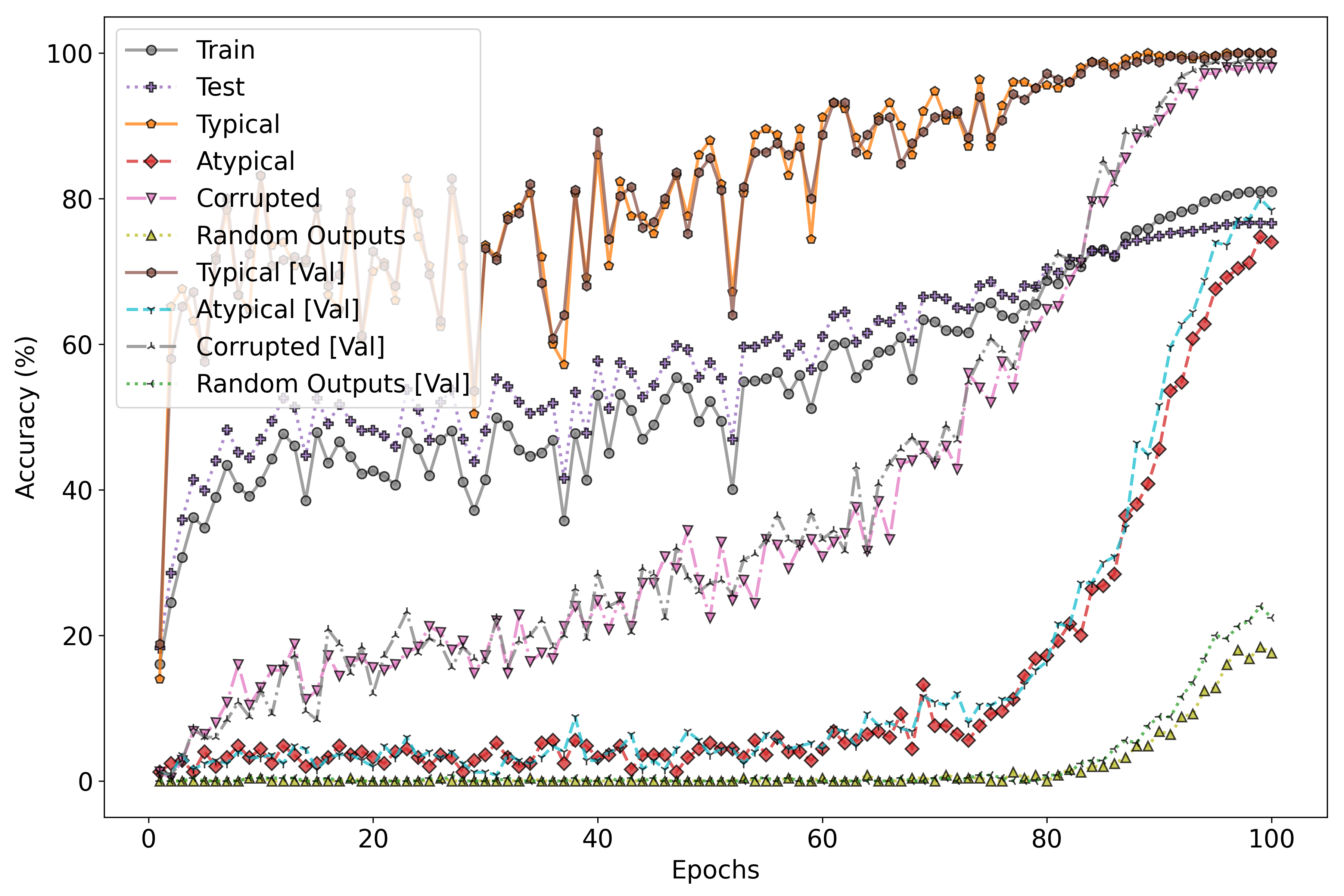}
    }
    \subfloat[Percent First-Learned]{
        \includegraphics[width=0.32\linewidth]{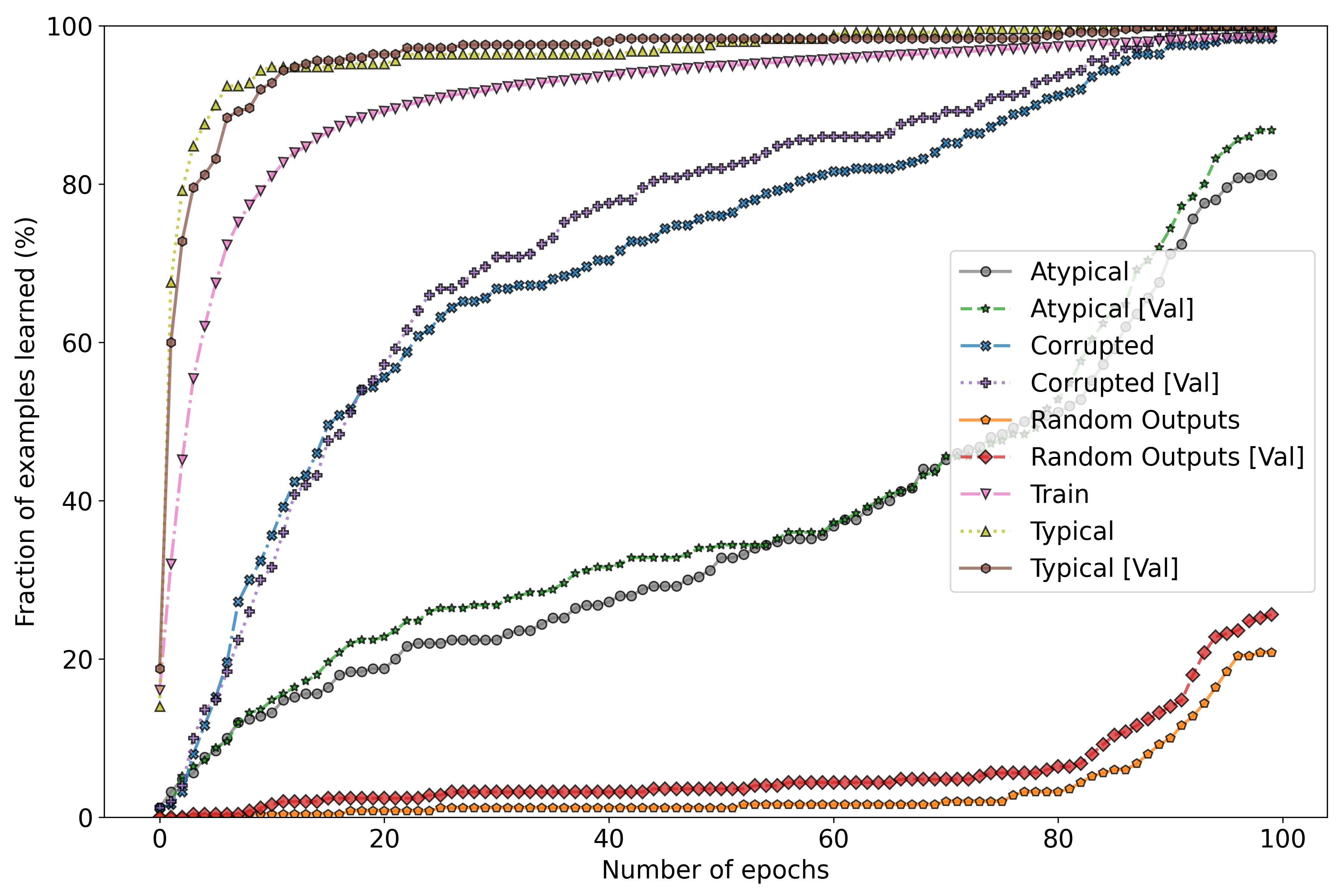}
    }
    \subfloat[Percent Consistently-Learned]{
        \includegraphics[width=0.32\linewidth]{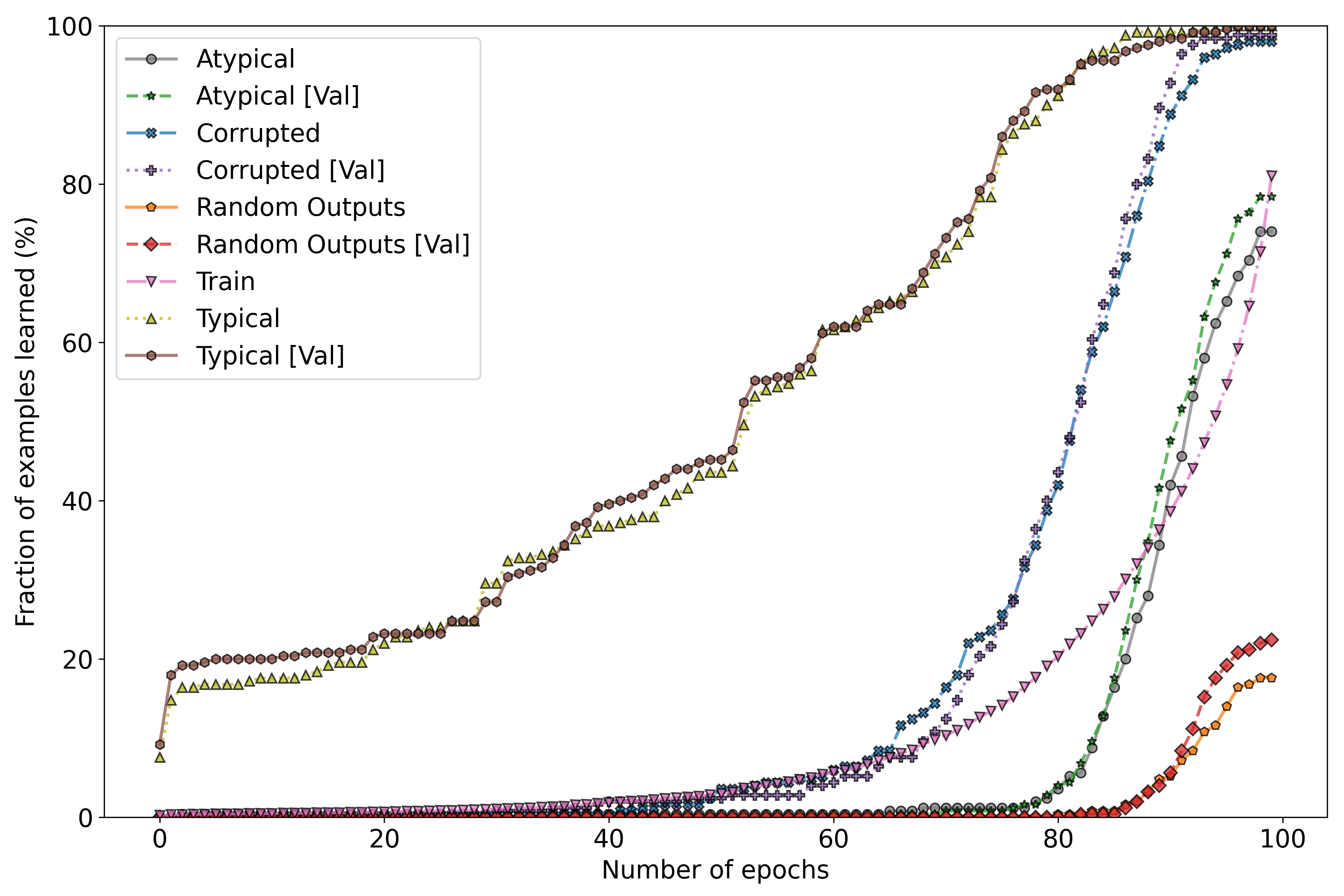}
    }
    
    \subfloat[1\textsuperscript{st} Epoch]{
        \includegraphics[width=0.18\linewidth]{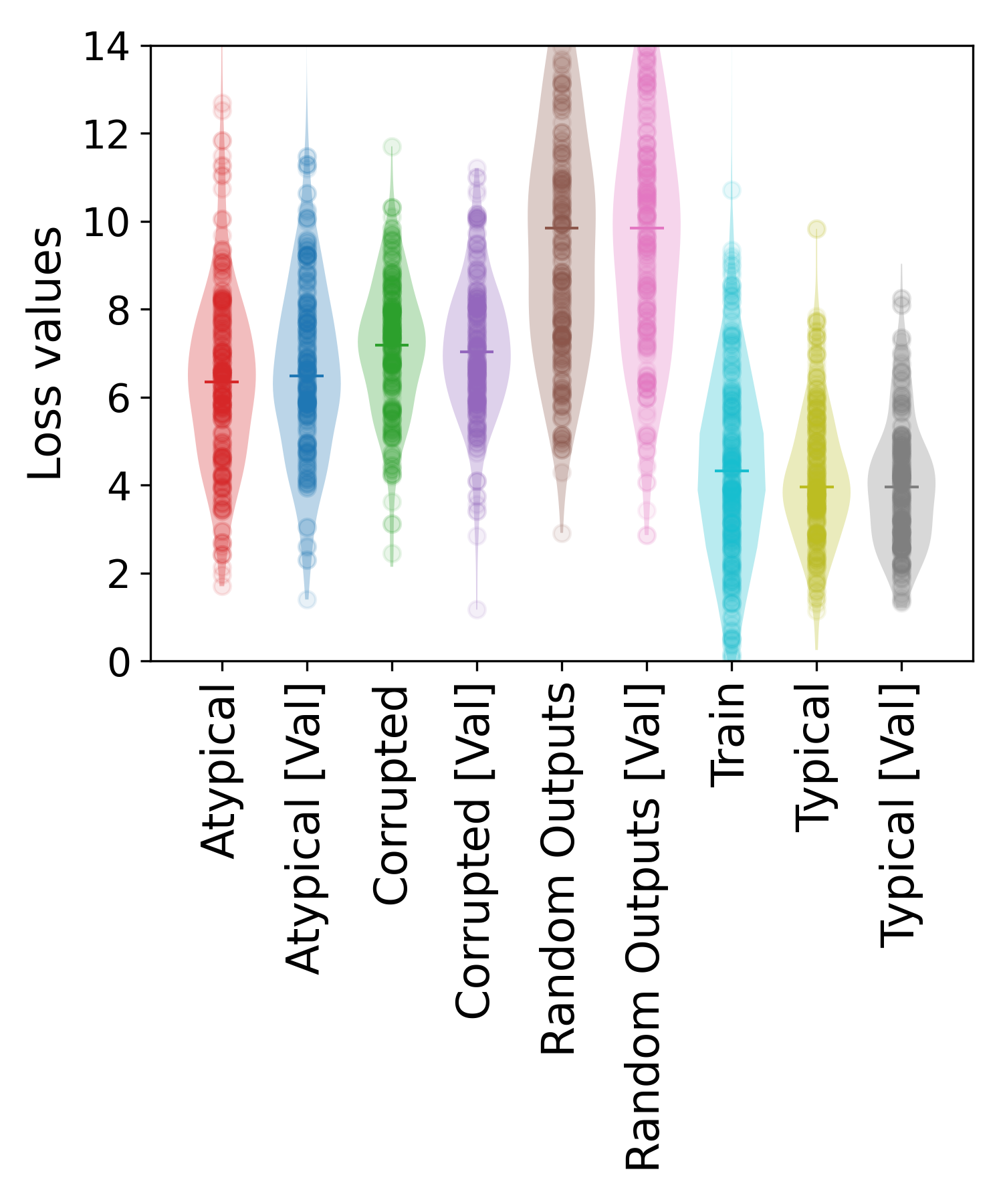}
    }
    \subfloat[25\textsuperscript{th} Epoch]{
        \includegraphics[width=0.18\linewidth]{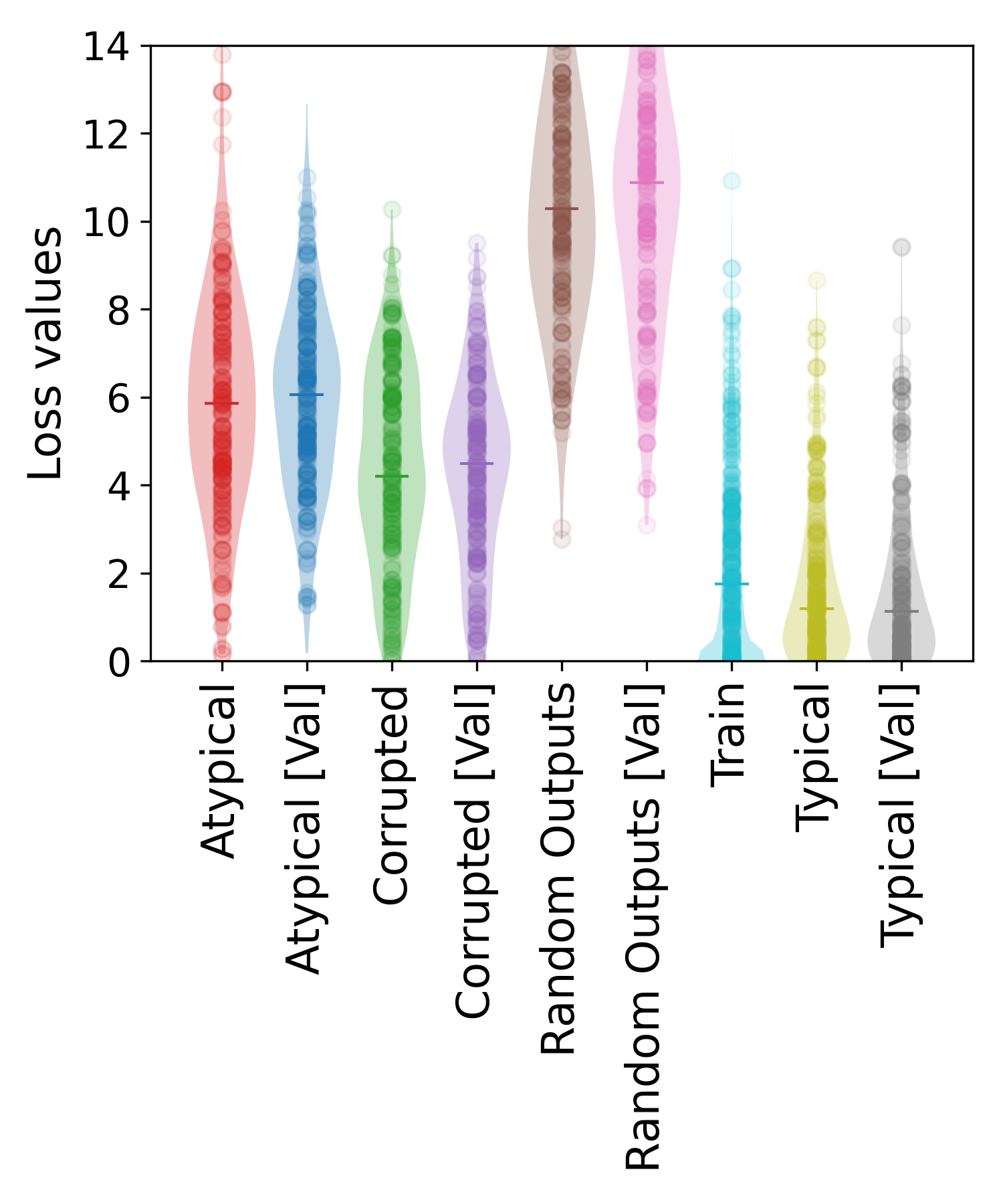}
    }
    \subfloat[50\textsuperscript{th} Epoch]{
        \includegraphics[width=0.18\linewidth]{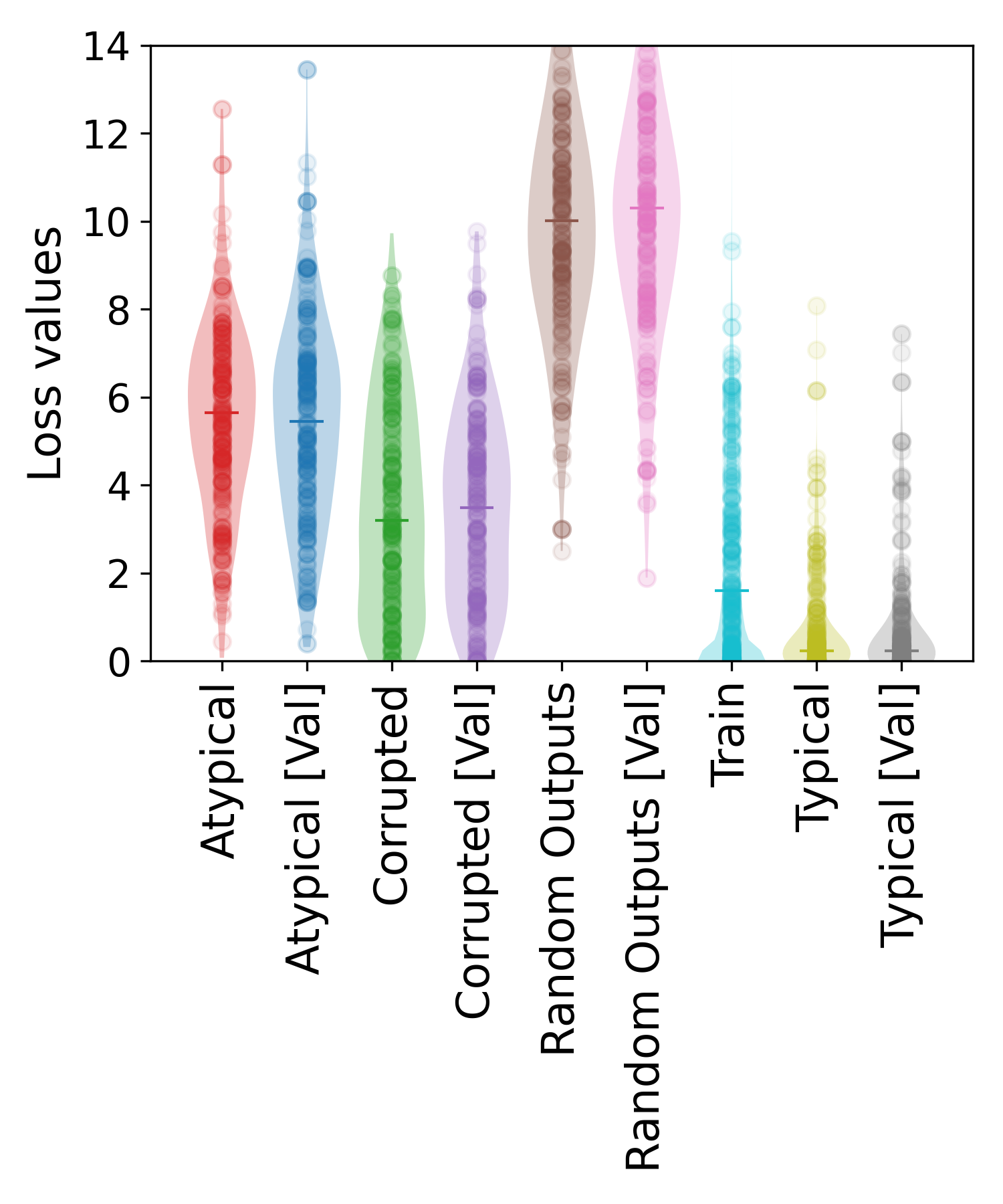}
    }
    \subfloat[75\textsuperscript{th} Epoch]{
        \includegraphics[width=0.18\linewidth]{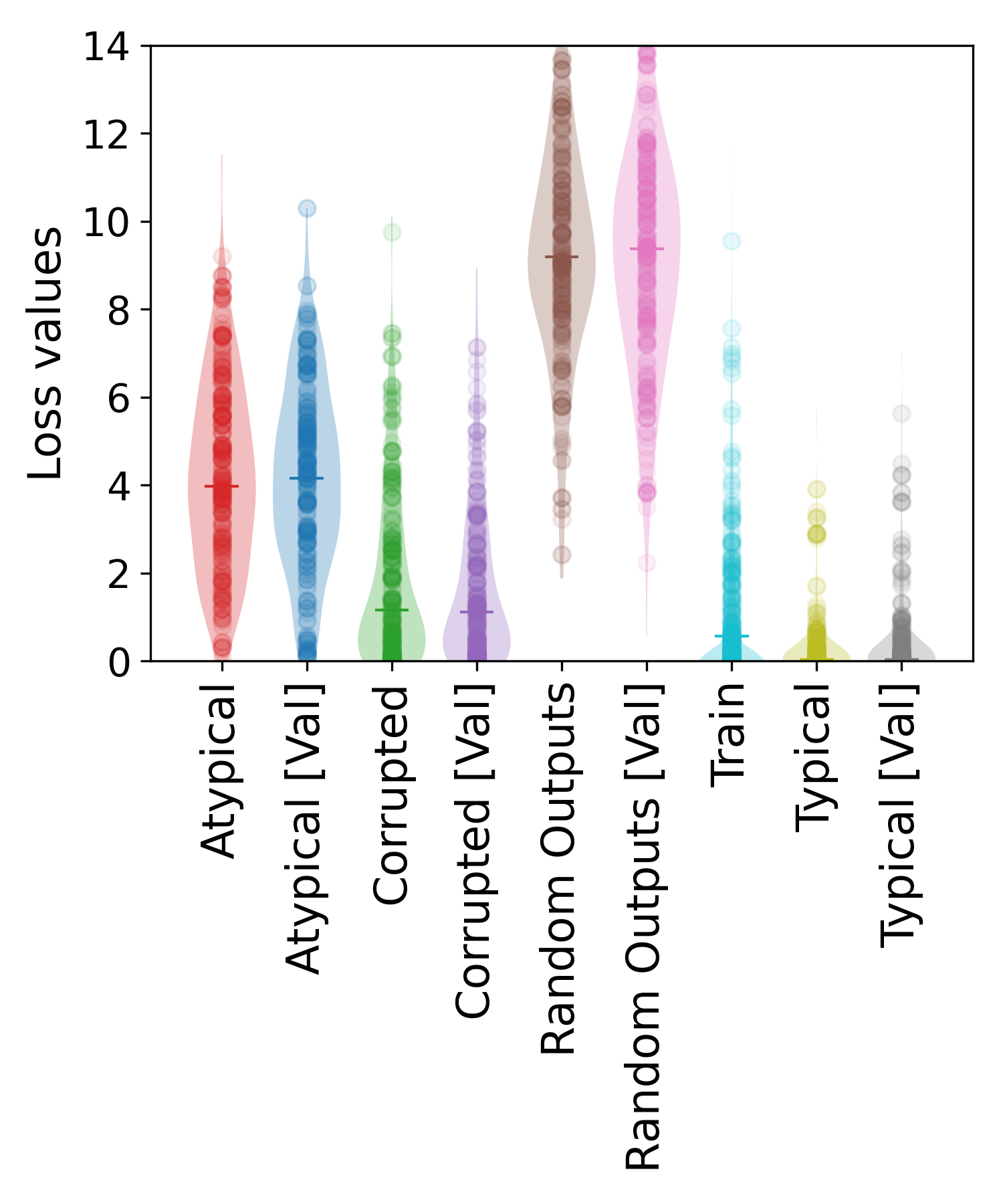}
    }
    \subfloat[90\textsuperscript{th} Epoch]{
        \includegraphics[width=0.18\linewidth]{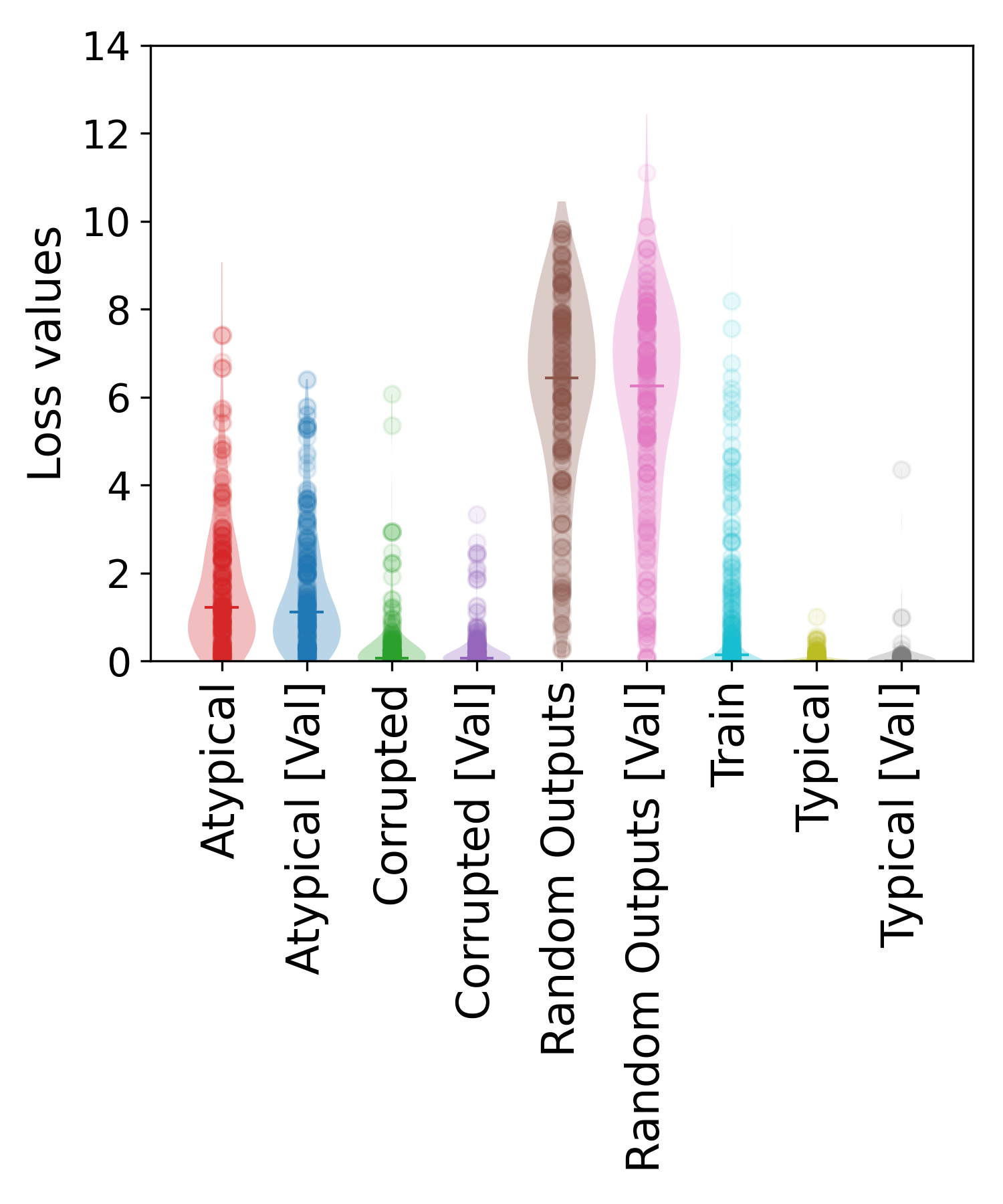}
    }
    
    \caption{%
    Probe categories are distinguishable via learning dynamics of a ResNet-50 trained on ImageNet, validating the approach of \textit{MAP-D}.
    For each of the probe categories and at each epoch, we plot  \textbf{(a)} each probe's average accuracy; \textbf{(b)} the cumulative fraction of examples once predicted correctly by the nth epoch;  and \textbf{(c)} the fraction that remain predicted correctly on all subsequent epochs.  Bottom plots \textbf{(d)-(h)} show the spread of losses at various epochs of training. 
    }
    \label{fig:training_dynamics_im_r50}
\end{figure*}

 \textit{MAP-D} works by comparing the performance of a given example to the learning curves typical of a given probe type. Our approach is motivated by the observation that different types of examples often exhibit very different learning dynamics over the course of training (see \Cref{fig:training_dynamics_im_r50}). In an empirical risk minimization setting, we minimize the average training loss across all training points.
$$
L_(\theta)=\frac{1}{N} \sum_{i=1}^{N} \ell\left(x_{i}, y_{i} ; \theta\right)
$$

However, performance on a given subset will differ from the average error.
Specifically, we firstly evaluate the learning curves of individual examples:
\begin{equation}
    \textbf{s}_{i}^{t} := \left( \ell(x_i, y_i ; \theta_{1}), \ell(x_i, y_i ; \theta_{2}), ..., \ell(x_i, y_i ; \theta_{t}) \mid (x_i, y_i) \in \mathcal{D} \right)
\end{equation}

\noindent where $\textbf{s}_{i}^{t}$ denotes the learning curve for the i\textsuperscript{th} training example, and $t$ is the current epoch\footnote{A coarser or finer resolution for the learning curves could also be used, e.g. every $n$ steps or epochs. All experiments in this work use differences computed at the end of the epoch.}.
We then track the per-example performance on probes \textbf{g} for each metadata category $m \in \{m_1, \dots, m_{|\mathcal{M}|}\}$, and refer to each probe as \textbf{g}($m$).

\begin{equation}
    \textbf{g}_{j}^{t}(m) := \left( \ell(x_j, y_j ; \theta_{1}), \ell(x_j, y_j ; \theta_{2}), ..., \ell(x_j, y_j ; \theta_{t}) \mid (x_j, y_j) \in \mathcal{D}_{m} \right)
\end{equation}

\noindent where $\textbf{g}_{j}^{t}(m)$ denotes the learning curve computed on the j\textsuperscript{th} example chosen from a given probe category $m$. We use $\mathcal{D}_{g}$ as shorthand to refer to the set of all these trajectories for the different probe categories along with the category identity.

\begin{equation}
    \mathcal{D}_{\textbf{g}} := \left( (\textbf{g}_{1}^{t}(m_1), m_1), \dots, (\textbf{g}_{|m_1|}^{t}(m_1), m_1), (\textbf{g}_{1}^{t}(m_2), m_2), \dots,  (\textbf{g}_{|m_{|\mathcal{M}|}|}^{t}(m_{|\mathcal{M}|}), m_{|\mathcal{M}|})\right)
    \label{eq:probe_dataset}
\end{equation}

\noindent where $|m_c|$ refers the number of examples belonging to the metadata category $m_c$.

We assign metadata features to an unseen data point by looking up the example's nearest neighbour from $\mathcal{D}_\textbf{g}$, using the Euclidean distance. In general, assignment of probe type could be done via any classification algorithm. However, in this work we use $k$-NN ($k$-Nearest Neighbours) for its simplicity, interpretability and the ability to compute the probability of multiple different metadata features. 

\begin{equation}
    p(m \mid \textbf{s}_{i}^{t}) = \frac{1}{k} \; \sum_{(\mathbf{g}, \hat{m}) \; \in \; \textsc{NN}(\textbf{s}_{i}^{t}, \mathcal{D}_\textbf{g}, k)} \mathbbm{1}_{\hat{m} = m} 
\end{equation}

\noindent where $p(m \mid \textbf{s}_{i}^{t})$ is the probability assigned to probe category $m$ based on the $k$ nearest neighbors for learning curve of the i\textsuperscript{th} training example from the dataset, and $\textsc{NN}(\textbf{s}_{i}^{t}, \mathcal{D}_g, k)$ represents the top-$k$ nearest neighbors for $\textbf{s}_{i}^{t}$ from $\mathcal{D}_\textbf{g}$ (probe trajectory dataset) based on Euclidean distance between the loss trajectories for all the probe examples and the given training example.  We fix $k$=20 in all our experiments.

This distribution over probes (i.e metadata features) may be of primary interest, but we are sometimes also interested in seeing which metadata feature a given example most strongly corresponds to; in this case, we compute the argmax:

\begin{equation}
    m'_i = \argmax_{m \in \mathcal{M}} \; p(m \mid \textbf{s}_{i}^{t})
\end{equation}

\noindent where $m'_i$ denotes the assignment of the i\textsuperscript{th} example to a particular probe category. 

We include the probe examples in the training set unless specified otherwise; excluding them in training can result in a drift in trajectories, and including them allows tracking of training dynamics. 

\begin{figure*}[!t]\centering
    \begin{minipage}[t]{\linewidth/6}\centering
        \strut Black bear\\
        \includegraphics[width=0.9\linewidth]{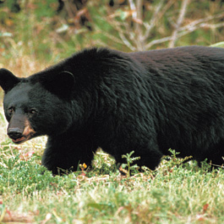}
        \subcaption{Typical}
    \end{minipage}\hfill
    \begin{minipage}[t]{\linewidth/6}\centering
        \strut Dishwasher\\
        \includegraphics[width=0.9\linewidth]{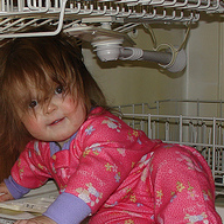}
        \subcaption{Atypical}\label{fig:probes_im_atypical}
    \end{minipage}\hfill
    \begin{minipage}[t]{\linewidth/6}\centering
        \strut School bus\\
        \includegraphics[width=0.9\linewidth]{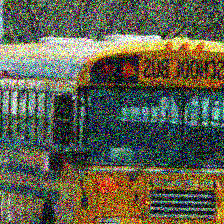}
        \subcaption{Corrupted}\label{fig:probes_im_corrupted}
    \end{minipage}\hfill
    \begin{minipage}[t]{\linewidth/6}\centering
        \strut Mud turtle\\
        \includegraphics[width=0.9\linewidth]{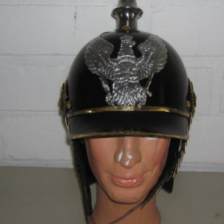}
        \subcaption{Rand Label}\label{fig:probes_im_mislabeled}
    \end{minipage}\hfill
    \begin{minipage}[t]{\linewidth/6}\centering
        \strut Jeep\\
        \includegraphics[width=0.9\linewidth]{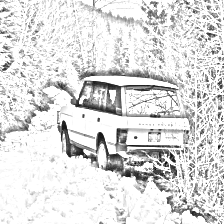}
        \subcaption{OOD}\label{fig:probes_im_ood_sketch}
    \end{minipage}\hfill
    \begin{minipage}[t]{\linewidth/6}\centering
        \strut Loafer\\
        \includegraphics[width=0.9\linewidth]{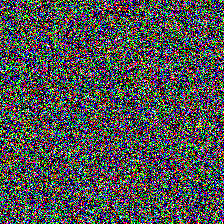}
        \subcaption{Rand Input}\label{fig:probes_noisy}
    \end{minipage}
    
    \caption{An illustration of samples from our curated probes. Creating our probe suites requires no human annotation. We curate different subsets of data that might exist in a dataset including (a) typical, (b) atypical, (c) corrupted, (d) mislabeled, (e) out-of-distribution, and (f) random input examples, using simple transformations or existing scoring mechanisms.}
    \label{fig:probes_imagenet}
\end{figure*}

\subsection{Probe Suite Curation}\label{sec:probe_curation}

While probe suites can be constructed using human annotations, this can be very expensive to annotate \citep{facct_mckane,veale_2017}. In many situations where auditing is desirable (e.g. toxic or unsafe content screening), extensive human labour is undesirable or even unethical \citep{Steiger2021ThePW,Shmueli2021BeyondFP}. 
Hence, in this work, we focus on probes that can be computationally constructed for arbitrary datasets -- largely by using simple transformations and little domain-specific knowledge. We emphasize that our probe suite is not meant to be exhaustive, but to provide enough variety in metadata features to demonstrate the merits of metadata archaeology.

\vspace{2mm}
\noindent We visualize these probes in \Cref{fig:probes_imagenet}, and describe below:

\begin{enumerate}
\itemsep0em
    \item \textbf{Typical} We quantify typicality by thresholding samples with the top consistency scores from~\citet{2020Jiang} across all datasets. The consistency score is a measure of expected classification performance on a held-out instance given training sets of varying size sampled from the training distribution.
    \item  \textbf{Atypical} Similarly, atypicality is quantified as samples with the lowest consistency scores from~\citet{2020Jiang}.
    \item \textbf{Random Labels} Examples in this probe have their labels replaced with uniform random labels, modelling label noise.
    \item \textbf{Random Inputs \& Labels} These noisy probes are comprised of uniform $\mathcal{U}(0, 1)$ noise sampled independently for every dimension of the input. We also randomly assign labels to these samples.
    \item \textbf{Corrupted Inputs} Corrupted examples are constructed by adding Gaussian noise with 0 mean and 0.1 standard deviation for CIFAR-10/100 and 0.25 standard deviation for ImageNet. These values were chosen to make the inputs as noisy as possible while still being (mostly) recognizable to humans.
\end{enumerate}

We curate 250 training examples for each probe category. For categories other than Typical/Atypical, we sample examples at random and then apply the corresponding transformations. All of the probes are then included during training so that we can study their training dynamics. We also curate 250 test examples for each probe category to evaluate the accuracy of our nearest neighbor assignment of metadata to unseen data points, where we know the true underlying metadata.

\section{Experiments and Discussion}

\begin{figure*}[!t]
    \centering
    \subfloat[Loss trajectories w/ four main probe categories]{
        \includegraphics[width=0.62\linewidth]{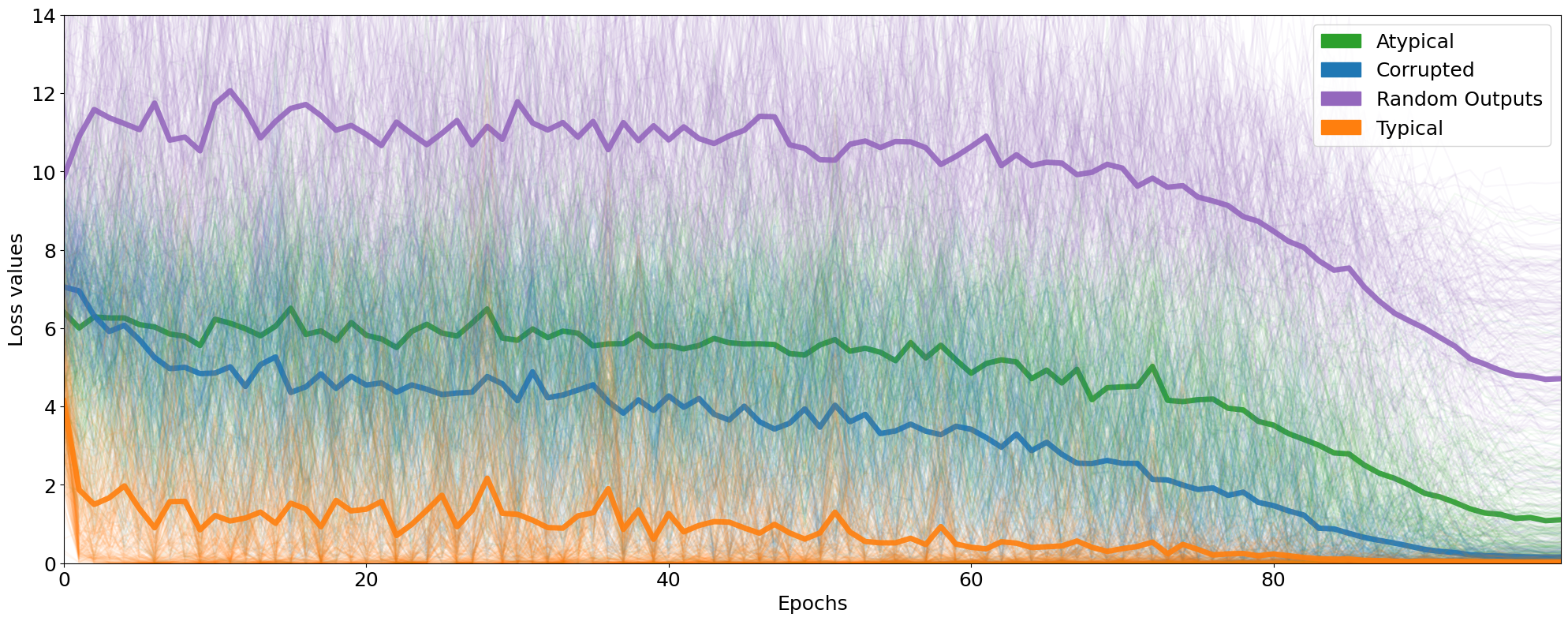}
    }
    \hspace{6mm}
    \subfloat[Confusion matrix]{
        \includegraphics[width=0.3\linewidth]{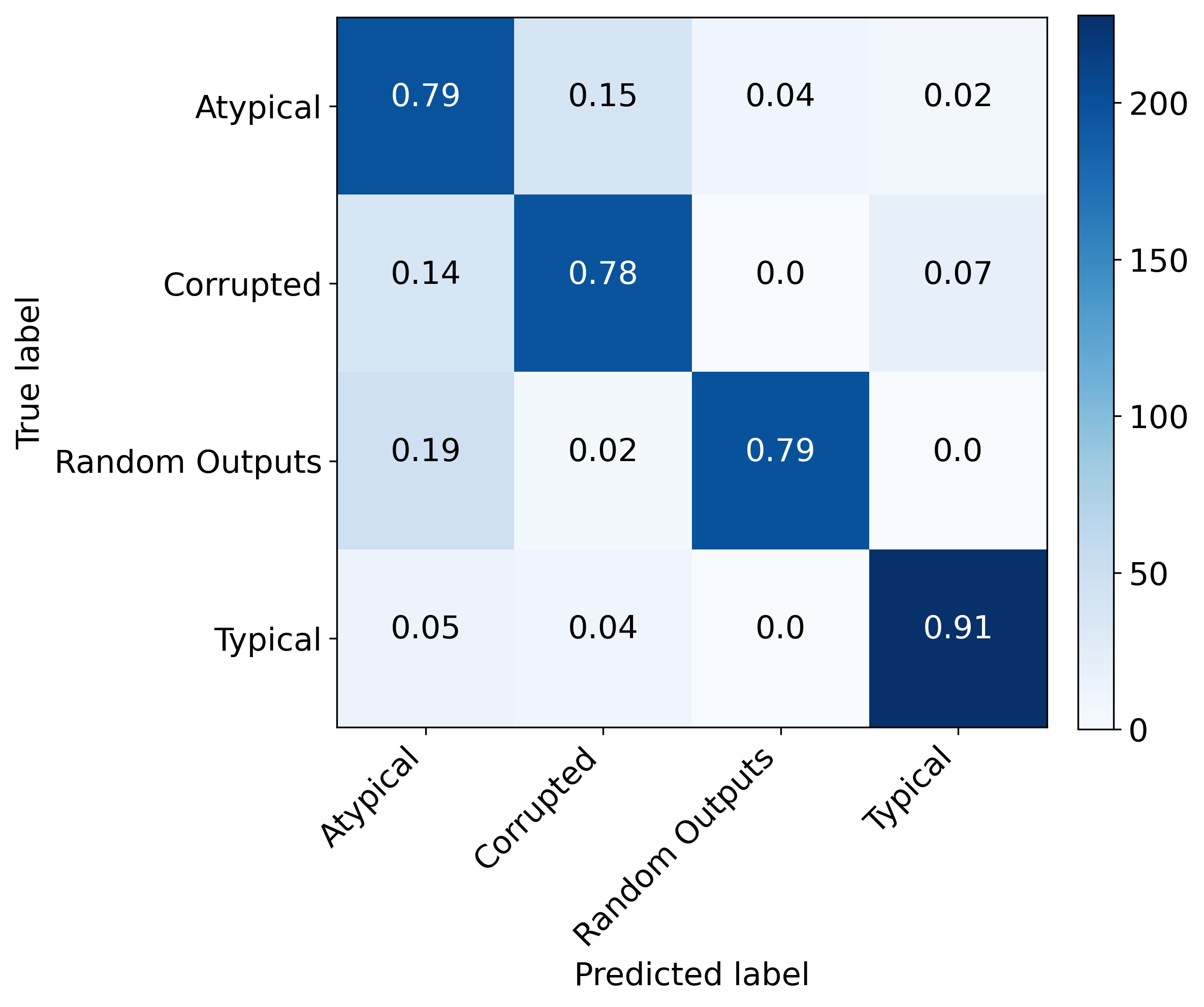}
    }

    \subfloat[Loss trajectories w/ all probe categories]{
        \includegraphics[width=0.62\linewidth]{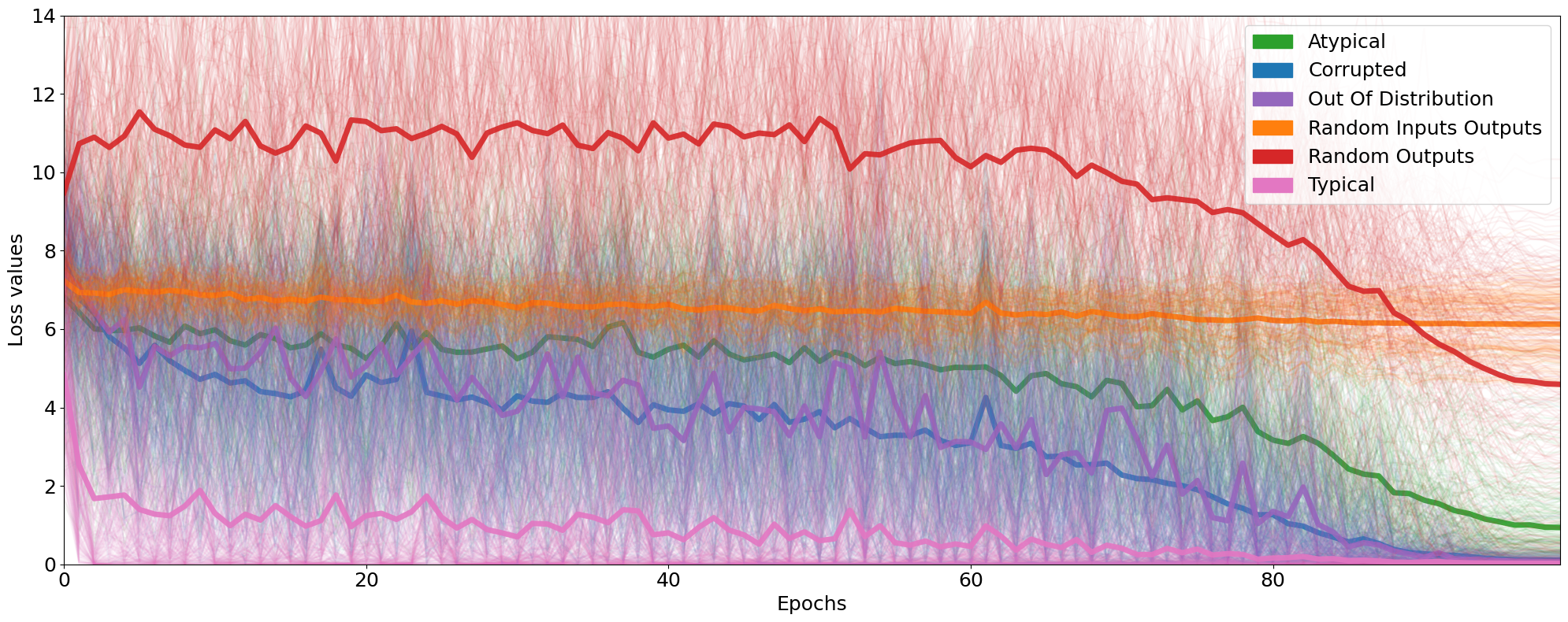}
    }
    \subfloat[Confusion matrix]{
        \includegraphics[width=0.38\linewidth]{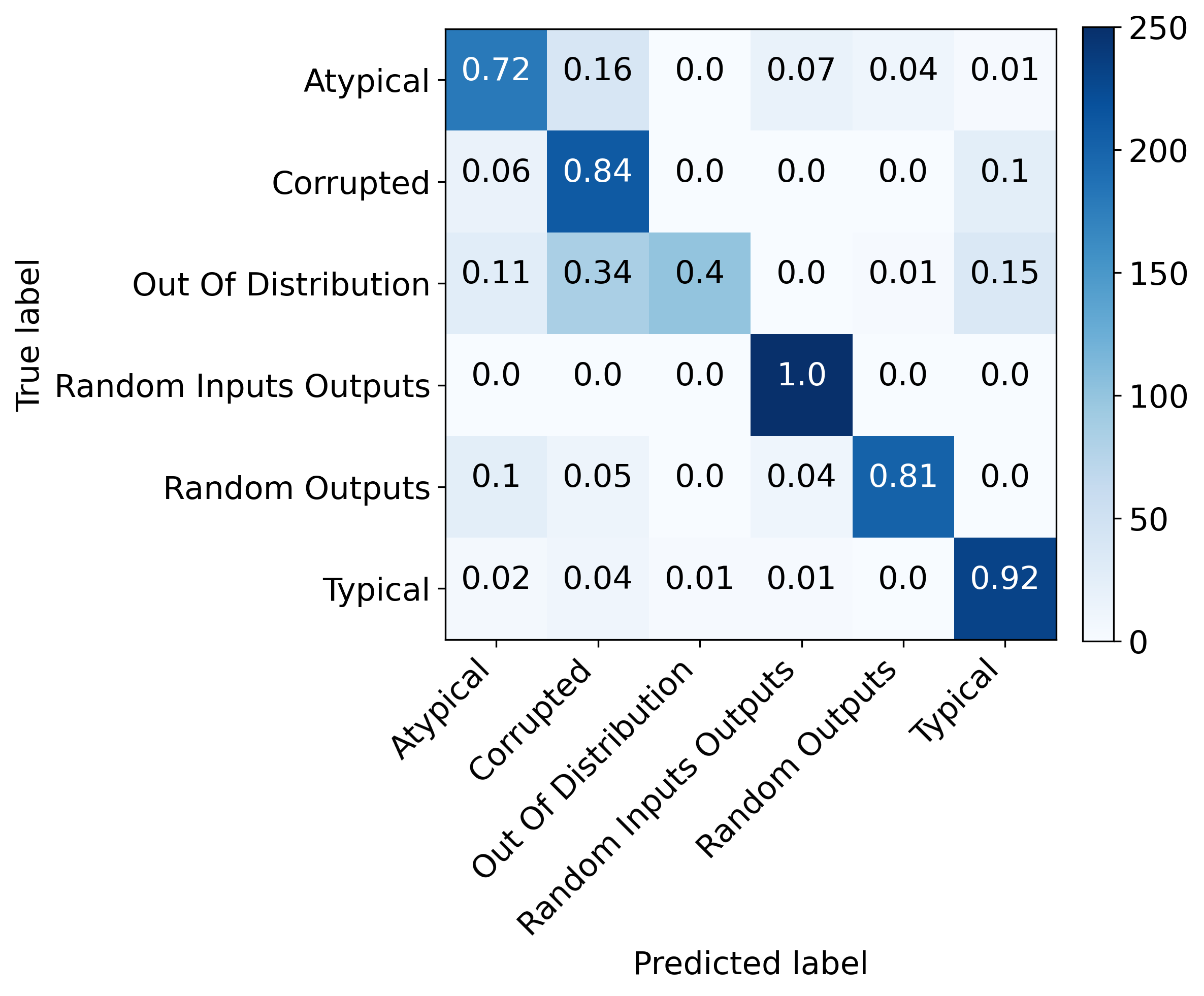}
    }
    
    \caption{Sanity check showing performance of \textit{MAP-D} on the probe suite test set using ResNet-50 on ImageNet, where we know the ground-truth metadata. \textbf{(a) and (c)} Solid line shows the mean learning curve while translucent lines are randomly sampled 250 individual trajectories for each probe category. Again, the separation of different probes is evident both in the dynamics over the course of training. \textbf{(b) and (d)} show confusion matrices between the true vs. predicted metadata features, demonstrating strong performance of the probes.}
    \label{fig:trajectory_classification_im_r50}
\end{figure*}

In the following sections, we perform experiments across 6 datasets: CIFAR-10/100, ImageNet, Waterbirds, CelebA, and Clothing1M. For details regarding the experimental setup, see~\Cref{sec:expdetails}. We first evaluate \textbf{convergence dynamics} of different probe suites (\Cref{sec:dynamics}), validating the approach of \textit{MAP-D}. We then qualitatively demonstrate the ability to \textbf{audit datasets} using \textit{MAP-D} (\Cref{sec:auditing}), and evaluate performance on a variety of downstream tasks: \textbf{noise correction }(\Cref{sec:label_correction}), \textbf{ prioritizing points for training} (\Cref{sec:prioritized_training}), and\textbf{ identifying minority-group samples} (\Cref{sec:minority_group}).

\subsection{Probe Suite Convergence Dynamics}\label{sec:dynamics}
In~\Cref{fig:training_dynamics_im_r50}, we present the training dynamics on the probe suites given a ResNet-50 model on ImageNet. For all datasets, we observe that probe suites have distinct learning convergence trajectories, demonstrating the efficacy of leveraging differences in training dynamics for the identification of probe categories.
We plot average 1) \textbf{Probe Accuracy} over the course of training, 2) the \textbf{Percent First-Learned} i.e. the percentage of samples which have been correctly classified once (even if that sample was be later forgotten) over the course of training, and 3) the \textbf{Percent Consistently-Learned} i.e. the percentage of samples which have been learned and will not be forgotten for the rest of training.
 
We observe consistent results across all dimensions. Across datasets, the \texttt{Typical} probe has the fastest rate of learning, whereas  the \texttt{Random Outputs} probe has the slowest.
When looking at \textit{Percent First-Learned} in~\Cref{fig:training_dynamics_im_r50}, we see a very clear natural sorting by the difficulty of different probes, where natural examples are learned earlier as compared to corrupted examples with synthetic noise. Examples with random outputs are the hardest for the model.

We also observe that probe ranking in terms of both \textit{Percent First-Learned} and \textit{Percent Consistently-Learned} is stable across training, indicating that model dynamics can be leveraged consistently as a stable signal to distinguish between different subsets of the distribution at any point in training. These results motivate our use of learning curves as signal to infer unseen metadata.

\subsection{Auditing Datasets}\label{sec:auditing}

\begin{figure*}[!t]\centering
    \begin{minipage}[t]{0.49\textwidth}\centering
        \begin{minipage}[t]{0.263\linewidth}
            \strut Typical \\
            \includegraphics[width=\linewidth,frame]{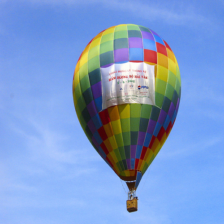} 
        \end{minipage}\hspace{3mm}
        \begin{minipage}[t]{0.3\linewidth}
            \strut \\
            \includegraphics[width=\linewidth]{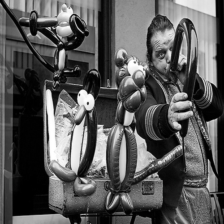}
        \end{minipage}
        \begin{minipage}[t]{0.3\linewidth}
            \strut \\
            \includegraphics[width=\linewidth]{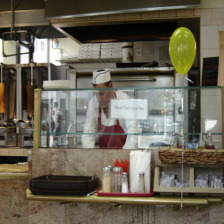}
        \end{minipage}
        \strut{Balloon}
    \end{minipage}\hfill
    \begin{minipage}[t]{0.49\textwidth}\centering
        \begin{minipage}[t]{0.263\linewidth}
            \strut Typical \\
            \includegraphics[width=\linewidth,frame]{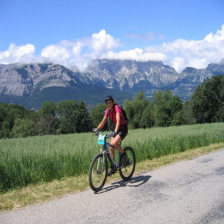}
        \end{minipage}\hspace{3mm}
        \begin{minipage}[t]{0.3\linewidth}
            \strut \\
            \includegraphics[width=\linewidth]{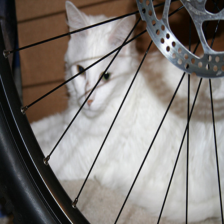}
        \end{minipage}
        \begin{minipage}[t]{0.3\linewidth}
            \strut \\
            \includegraphics[width=\linewidth]{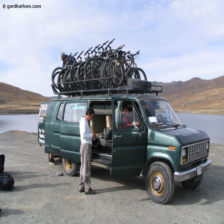}
        \end{minipage}
        \strut{Mountain Bike}
    \end{minipage}

    \begin{minipage}[t]{0.49\textwidth}\centering
        \begin{minipage}[t]{0.263\linewidth}
            \strut Typical \\
            \includegraphics[width=\linewidth,frame]{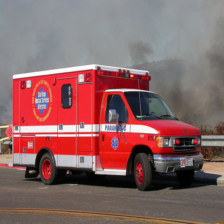}
        \end{minipage}\hspace{3mm}
        \begin{minipage}[t]{0.3\linewidth}
            \strut \\
            \includegraphics[width=\linewidth]{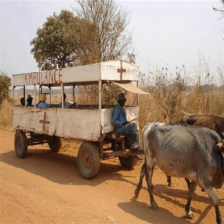}
        \end{minipage}
        \begin{minipage}[t]{0.3\linewidth}
            \strut \\
            \includegraphics[width=\linewidth]{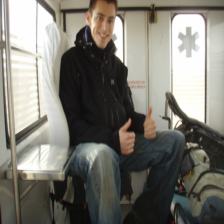}
        \end{minipage}
        \strut{Ambulance}
    \end{minipage}
    \begin{minipage}[t]{0.49\textwidth}\centering
        \begin{minipage}[t]{0.263\linewidth}
            \strut Typical \\
            \includegraphics[width=\linewidth,frame]{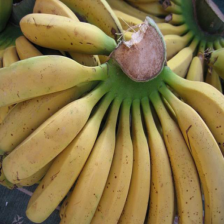}
        \end{minipage}\hspace{3mm}
        \begin{minipage}[t]{0.3\linewidth}
            \strut \\
            \includegraphics[width=\linewidth]{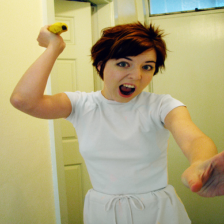}
        \end{minipage}
        \begin{minipage}[t]{0.3\linewidth}
            \strut \\
            \includegraphics[width=\linewidth]{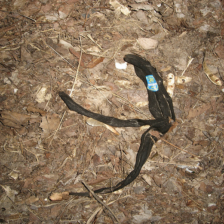}
        \end{minipage}
        \strut{Banana}
    \end{minipage}

    \begin{minipage}[t]{0.49\textwidth}\centering
        \begin{minipage}[t]{0.263\linewidth}
            \strut Typical \\
            \includegraphics[width=\linewidth,frame]{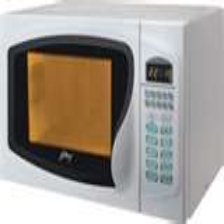}
        \end{minipage}\hspace{3mm}
        \begin{minipage}[t]{0.3\linewidth}
            \strut \\
            \includegraphics[width=\linewidth]{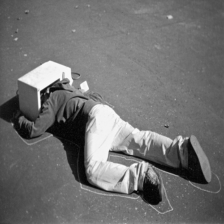}
        \end{minipage}
        \begin{minipage}[t]{0.3\linewidth}
            \strut \\
            \includegraphics[width=\linewidth]{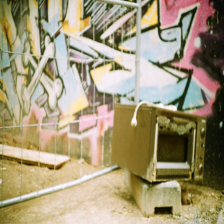}
        \end{minipage}
        \strut{Microwave}
    \end{minipage}
    \begin{minipage}[t]{0.49\textwidth}\centering
        \begin{minipage}[t]{0.263\linewidth}
            \strut Typical \\
            \includegraphics[width=\linewidth,frame]{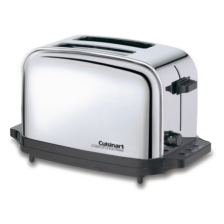}
        \end{minipage}\hspace{3mm}
        \begin{minipage}[t]{0.3\linewidth}
            \strut \\
            \includegraphics[width=\linewidth]{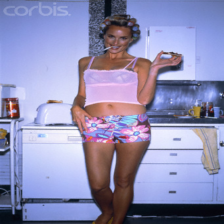}
        \end{minipage}
        \begin{minipage}[t]{0.3\linewidth}
            \strut \\
            \includegraphics[width=\linewidth]{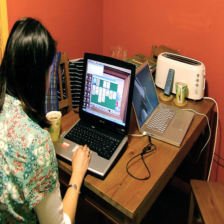}
        \end{minipage}
        \strut{Toaster}
    \end{minipage}

    \begin{minipage}[t]{0.49\textwidth}\centering
        \begin{minipage}[t]{0.263\linewidth}
            \strut Typical \\
            \includegraphics[width=\linewidth,frame]{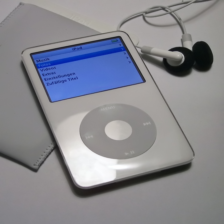}
        \end{minipage}\hspace{3mm}
        \begin{minipage}[t]{0.3\linewidth}
            \strut \\
            \includegraphics[width=\linewidth]{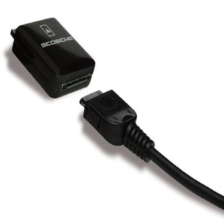}
        \end{minipage}
        \begin{minipage}[t]{0.3\linewidth}
            \strut \\
            \includegraphics[width=\linewidth]{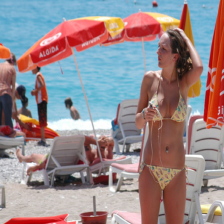}
        \end{minipage}
        \strut{iPod}
    \end{minipage}
    \begin{minipage}[t]{0.49\textwidth}\centering
        \begin{minipage}[t]{0.263\linewidth}
            \strut Typical \\
            \includegraphics[width=\linewidth,frame]{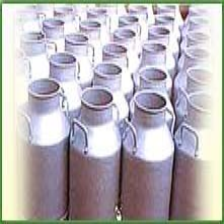}
        \end{minipage}\hspace{3mm}
        \begin{minipage}[t]{0.3\linewidth}
            \strut \\
            \includegraphics[width=\linewidth]{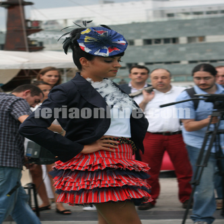}
        \end{minipage}
        \begin{minipage}[t]{0.3\linewidth}
            \strut \\
            \includegraphics[width=\linewidth]{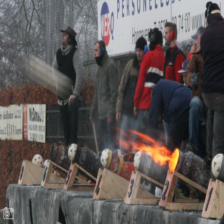}
        \end{minipage}
        \strut{Milk Can}
    \end{minipage}
    
    \caption{Examples surfaced through the use of \textit{MAP-D} on ImageNet train set using the \texttt{Typical} probe (first image in each set, highlighted with \textbf{black border}) and \texttt{Random Output} probe (next two images). Sub-caption indicates the ground truth class. This showcases the utility of \textit{MAP-D} for exploring a dataset, showing what the model considers typical for a class as well as  uncovering potentially problematic examples.}
    \label{fig:surfaced_examples_imagenet_random_outputs}
\end{figure*}
A key motivation of our work is that the large size of modern datasets means only a small fraction of datapoints can be economically inspected by humans.  In safety-critical or otherwise sensitive domains such as health care diagnostics \citep{2019Hongtao, Gruetzemacher20183DDL, 2019badgeley,2019oakden}, self-driving cars \citep{2017Telsa}, hiring \citep{Dastin_2018, Harwell_2019}, and many others, providing tools for domain experts to audit models is of great importance to ensure scalable oversight.

We apply \textit{MAP-D} to infer the metadata features of the underlying dataset. In Fig.~\ref{fig:surfaced_examples_imagenet}, we visualize class specific examples surfaced by \textit{MAP-D} on the ImageNet train set. Our visualization shows that \textit{MAP-D} helps to disambiguate effectively between different types of examples and can be used to narrow down the set of datapoints to prioritize for inspection.
We observe clear semantic differences between the sets. In Fig.~\ref{fig:surfaced_examples_imagenet}, we observe that examples surfaced as \texttt{Typical} are mostly well-centered images with a typical color scheme, where the only object in the image is the object of interest. Examples surfaced as \texttt{Atypical} present the object in unusual settings or vantage points, or feature differences in color scheme from the typical variants. We observe examples that would be hard for a human to classify using the \texttt{Random Output} probe category. For example, we see incorrectly labeled images of a digital watch, images where the labeled object is hardly visible, artistic and ambiguous images, and multi-object examples where several different labels may be appropriate. We visualize more examples from the \texttt{Random Output} probe category in Fig.~\ref{fig:surfaced_examples_imagenet_random_outputs}.

As a sanity check, in Fig.~\ref{fig:trajectory_classification_im_r50}  we also evaluate the performance of \textit{MAP-D} on the held-out probe test set,  where we know the true underlying metadata used to curate that example. In \textbf{(a)}, we compute performance on the four probes which are most easily separable via learning curves, and find that model was able to achieve high detection performance ($\sim 81.9\%$ accuracy). When including \texttt{OOD} and \texttt{Random Output} probe categories (bottom row \textbf{(c)} and \textbf{(d)}), we observe from the learning curves that there is overlap in probe dynamics; these two categories are difficult to disambiguate, and as a result, we see a slight drop in overall performance ($\sim 78\%$).

\subsection{Label Noise Correction} \label{sec:label_correction}

\begin{figure*}[!t]
    \centering
    \subfloat[CIFAR-10]{
        \includegraphics[width=0.48\linewidth]{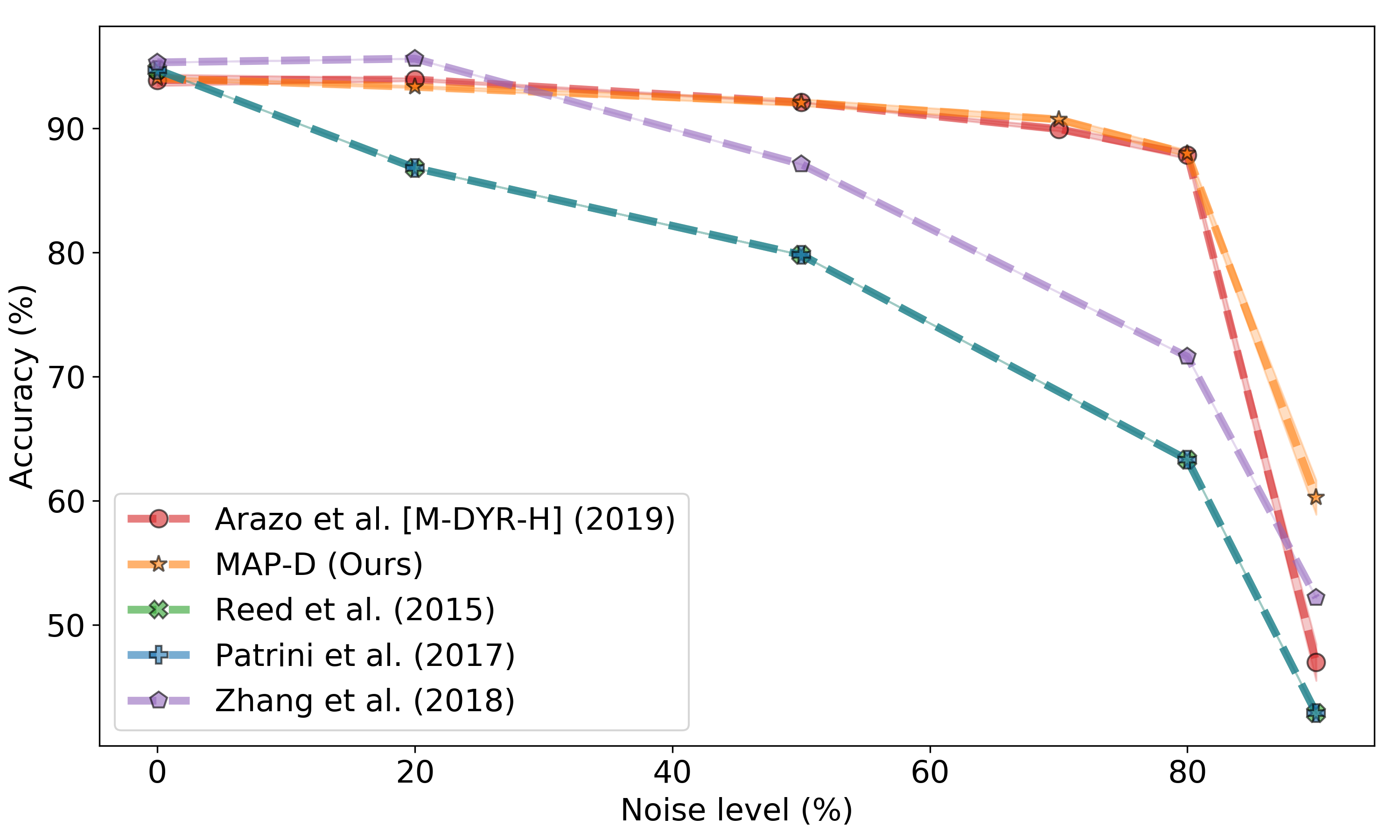}
    }
    \subfloat[CIFAR-100]{
        \includegraphics[width=0.48\linewidth]{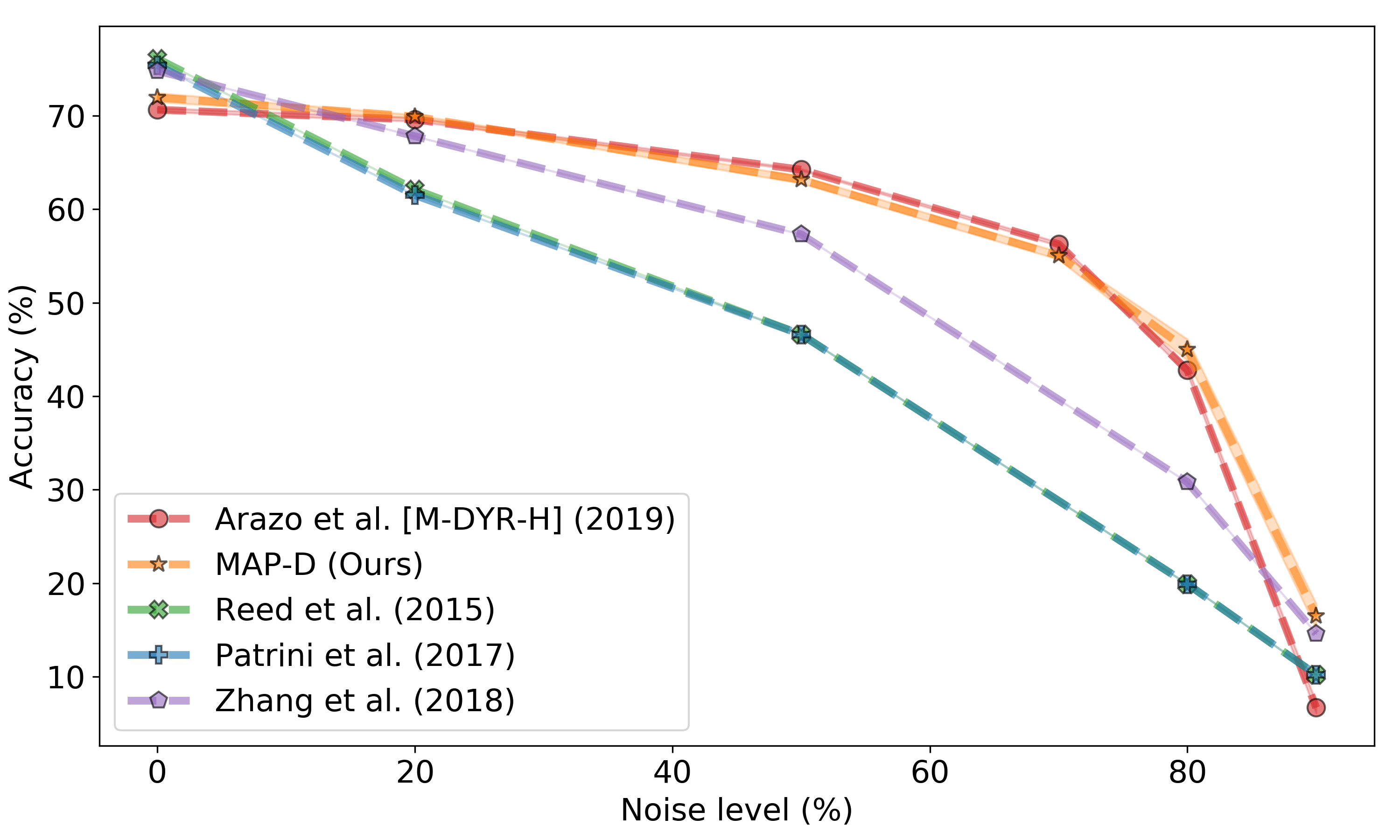}
    }
    \caption{Comparison of different noise correction methods under the presence of label noise. Mean and standard deviation reported over 3 random runs. \textit{MAP-D} is competitive with most other methods, many of which are particularly targeted towards this problem. }
    \label{fig:label_correction}
\end{figure*}

Here we apply \textit{MAP-D} to detect and correct label noise, a data quality issue that has been heavily studied in prior works~\cite{zhang2017mixup,arazo2019unsupervised,arpit2017closer}. We benchmark against a series of different baselines (\citet{arazo2019unsupervised,zhang2017mixup,patrini2017making,reed2014training}), some of which are specifically developed to deal with label noise. We emphasize that our aim is not to develop a specialized technique for dealing with label noise, but to showcase that \textit{MAP-D}, a general solution for metadata archaeology, also performs well on specialized tasks such as label correction.

To distinguish between clean and noisy samples using \textit{MAP-D}, we add an additional \textit{random sample} probe curated via a random sample from the (unmodified) underlying data, as a proxy for clean data. For this comparison, we follow the same experimental protocol as in~\cite{arazo2019unsupervised} where all the methods we benchmark against are evaluated.

Concretely, for any label correction scheme, the actual label used for training is a convex combination of the original label and the model's prediction based on the probability of the sample being either clean or noisy.
Considering one-hot vectors, the correct label can be represented as:

\begin{equation}
    \bar{y_{i}} = p(\texttt{clean} \mid \textbf{s}_{i}^{t}) \times y_i + p(\texttt{noisy} \mid \textbf{s}_{i}^{t}) \times \hat{y}_i
\end{equation}

\noindent where $\bar{y_{i}}$ represents the corrected label used to train the model, $y_i$ represents the label present in the dataset weighted by the probability of the sample being clean $p(\texttt{clean} \mid \textbf{s}_{i}^{t})$, and $\hat{y}_i$ represents the model's prediction (a one-hot vector computed via argmax rather than predicted probabilities) weighted by the probability of the sample being noisy $p(\texttt{noisy} \mid \textbf{s}_{i}^{t})$.
Since we are only considering two classes, $p(\texttt{clean} \mid \textbf{s}_i^t)=1-p(\texttt{noisy} \mid \textbf{s}_i^t)$.
We employ the online \textit{MAP-D} trajectory scheme in this case, where the learning curve is computed given all prior epochs completed as of that point.

Despite the relative simplicity and generality of \textit{MAP-D}, it generally performs as well as highly-engineereed methods developed specifically for this task. Our results are presented in Fig.~\ref{fig:label_correction}. Specifically, at  extremely \textbf{high levels of noise}, \textit{MAP-D} performs significantly better on both CIFAR-10 and CIFAR-100 as compared to Arazo et al.~\cite{arazo2019unsupervised} (CIFAR-10: $\sim 47\%$ vs $\sim 59\%$; CIFAR-100: $\sim 6.5\%$ vs $\sim 16.5\%$).

\begin{figure*}[!t]
    \centering
    \subfloat[CIFAR-10]{
        \includegraphics[width=0.48\linewidth]{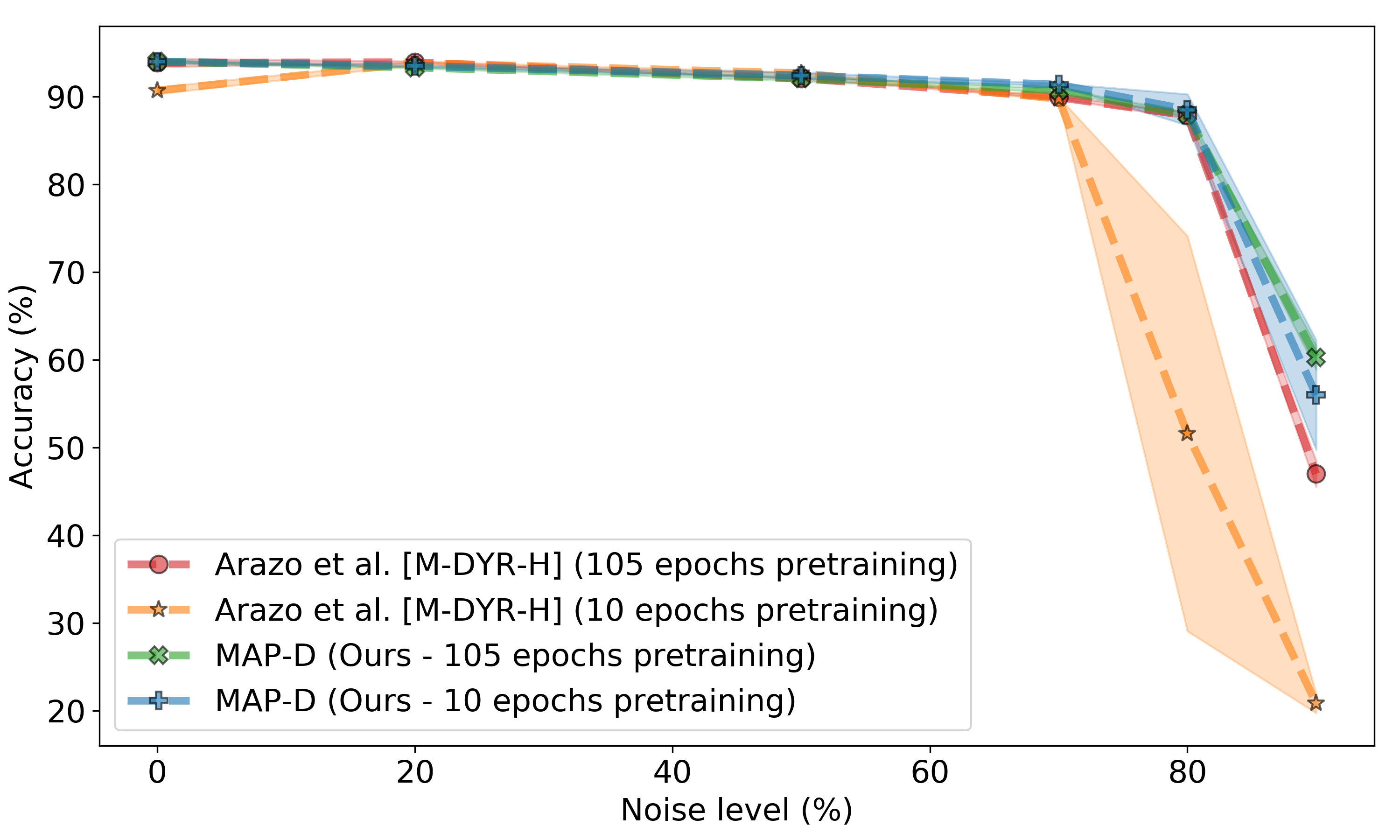}
    }
    \subfloat[CIFAR-100]{
        \includegraphics[width=0.48\linewidth]{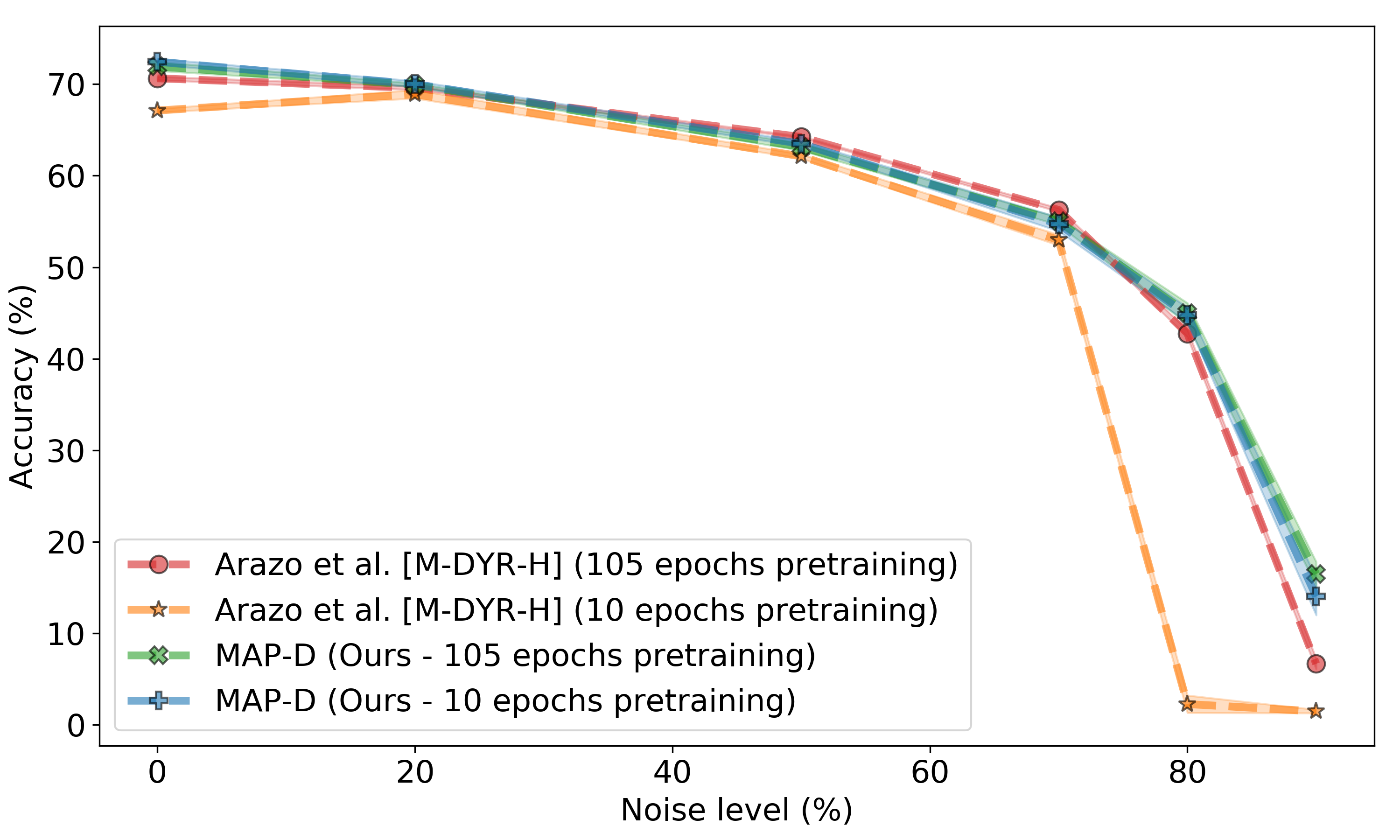}
    }
    \caption{Comparison between pretraining schedules of 105 epochs (default value as set by Arazo et al.~\cite{arazo2019unsupervised}) and 10 epochs. Mean and standard deviation reported over 3 random runs. \textit{MAP-D} is robust against changes in the number of pretraining epochs, while the method in~\cite{arazo2019unsupervised} achieves slightly poorer performance in the low-noise setting and  significantly  poorer performance in the high-noise setting.}
    \label{fig:label_correction_10ep_pretraining}
\end{figure*}

\paragraph{Number of epochs before noise correction} In the original setting proposed by ~\citep{arazo2019unsupervised}, the method requires pretraining for 105 epochs prior to label correction. We observe that a relative strength of \textit{MAP-D} is the ability to forgo such prolonged pretraining while retraining for robust noisy example detection. We perform a simple experiment with a reduced number of pretraining epochs (10 instead of 105), with results presented in Fig.~\ref{fig:label_correction_10ep_pretraining}, demonstrating that there is only negligible impact of pretraining schedule on \textit{MAP-D} performance, while  the performance of Arazo et al. (2019)~\cite{arazo2019unsupervised} is drastically impacted, specifically in no-noise and high-noise regimes.

\subsection{Prioritized Training} \label{sec:prioritized_training}

\begin{figure*}[!t]
    \centering
    \includegraphics[width=0.9\linewidth]{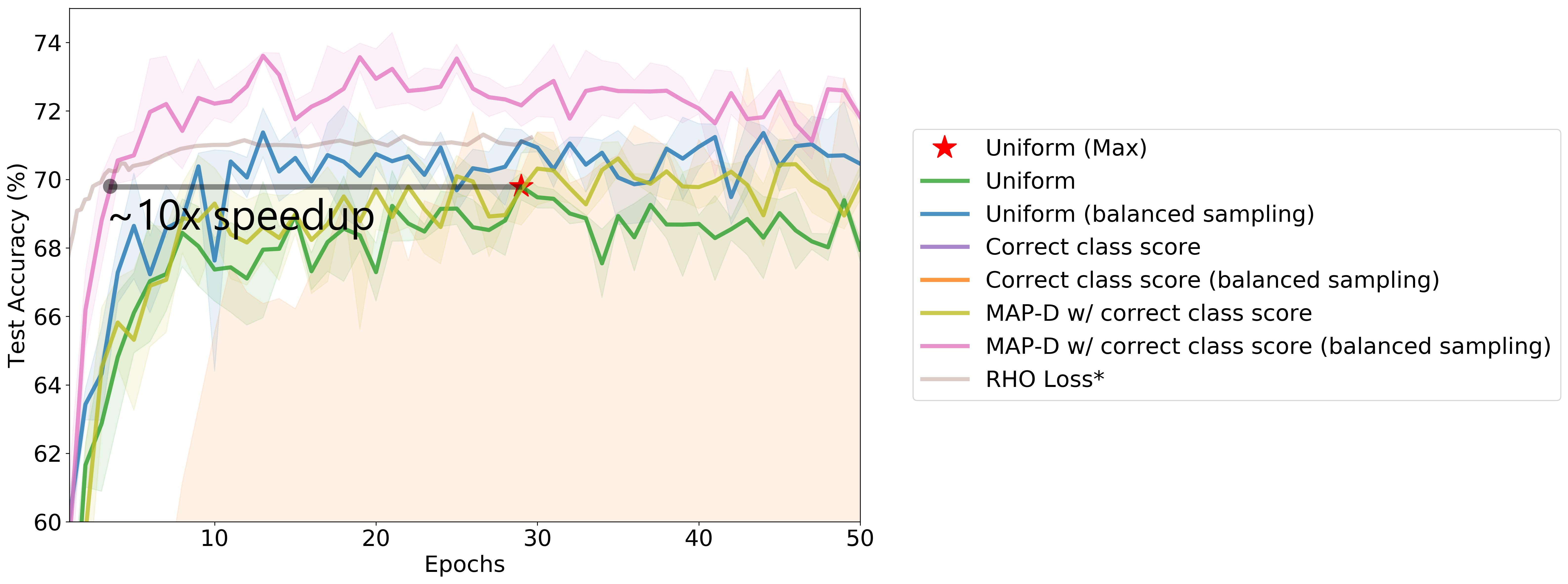}
    \caption{Results for score-based prioritization with \textit{MAP-D} (\textbf{pink, lavender}) compared against simple baselines. Mean and standard deviation computed over 3 random runs.
    The correct class score baselines (\textbf{purple, orange}) both select examples with the highest loss, which lead to poor performance due to label noise.
    Uniform selection baselines (\textbf{blue, green}) perform quite well, but take longer to train; out method achieves almost a 10x training speedup. RHO Loss* baseline (\textbf{grey))} plots original results reported in~\cite{mindermann2022prioritized}  while all other results use our implementation. While * use a different reporting interval, results remain comparable. The two methods are similar in training speed, but \textit{MAP-D} achieves higher accuracy.}
    \label{fig:prioritized_clothing_1m}
\end{figure*}

Prioritized training refers to selection of most useful points for training in an online fashion with the aim of speeding up the training process.
We consider the online batch selection scenario presented in~\cite{mindermann2022prioritized}, where we only train on a selected 10\% of the examples in each minibatch.
Simple baselines for this task include selecting points with high loss or at uniform random.
It can be helpful to prioritize examples which are not yet learned (i.e. consistently correctly classified), but this can also select for mislabeled examples, which are common in large web-scraped datasets such as Clothing1M~\cite{xiao2015clothing1m}.
As noted by \citet{mindermann2022prioritized}, we need to find points which are \textit{useful} to learn.
Applying \textit{MAP-D} in this context allows us to leverage training dynamics to identify such examples -  we look for examples that are not already learned, but which still have training dynamics that resemble clean data:
\begin{equation}
    \texttt{training\_score = (clean\_score + (1. - correct\_class\_confidence)) / 2.}
\end{equation}
\noindent where $\texttt{clean\_score}$ is the probability of an example being clean (vs. noisy) according to the k-NN classifier described in Section~\ref{subsec:assigning_examples}. 
An example can achieve a maximum score of 1 under this metric when\textit{ MAP-D} predicts the example is clean, but the model assigns 0 probability to the correct label.
Following \citet{mindermann2022prioritized}, we select 32 examples from each minibatch of 320.
For (class-)balanced sampling, we also ensure that we always select at least 2 examples from each of the 14 possible classes, which significantly improves performance.
Figure~\ref{fig:prioritized_clothing_1m} shows the effectiveness of this approach vs.\ these baselines; we achieve a 10x speedup over uniform random selection of examples.

We also report the original results from~\cite{mindermann2022prioritized} for reference which uses a different reporting interval. 
~\cite{mindermann2022prioritized} requires pretraining a separate model, and uses the prediction of that model to decide which points to prioritize for training.
Our method on the other hand uses an online \textit{MAP-D} trajectory scheme to decide whether an example is clean or noisy\footnote{We append the loss values of all examples in the batch to their learning curves before computing the assignments in order to ensure that examples can be correctly assigned even at the first epoch.}.
It is important to note that using balanced sampling with RHO Loss~\cite{mindermann2022prioritized} is likely to also improve performance for~\cite{mindermann2022prioritized}.

\subsection{Detection of Minority Group Samples} \label{sec:minority_group}

\begin{figure*}[!t]
    \centering
    \subfloat[Waterbirds]{
        \includegraphics[width=0.4\linewidth]{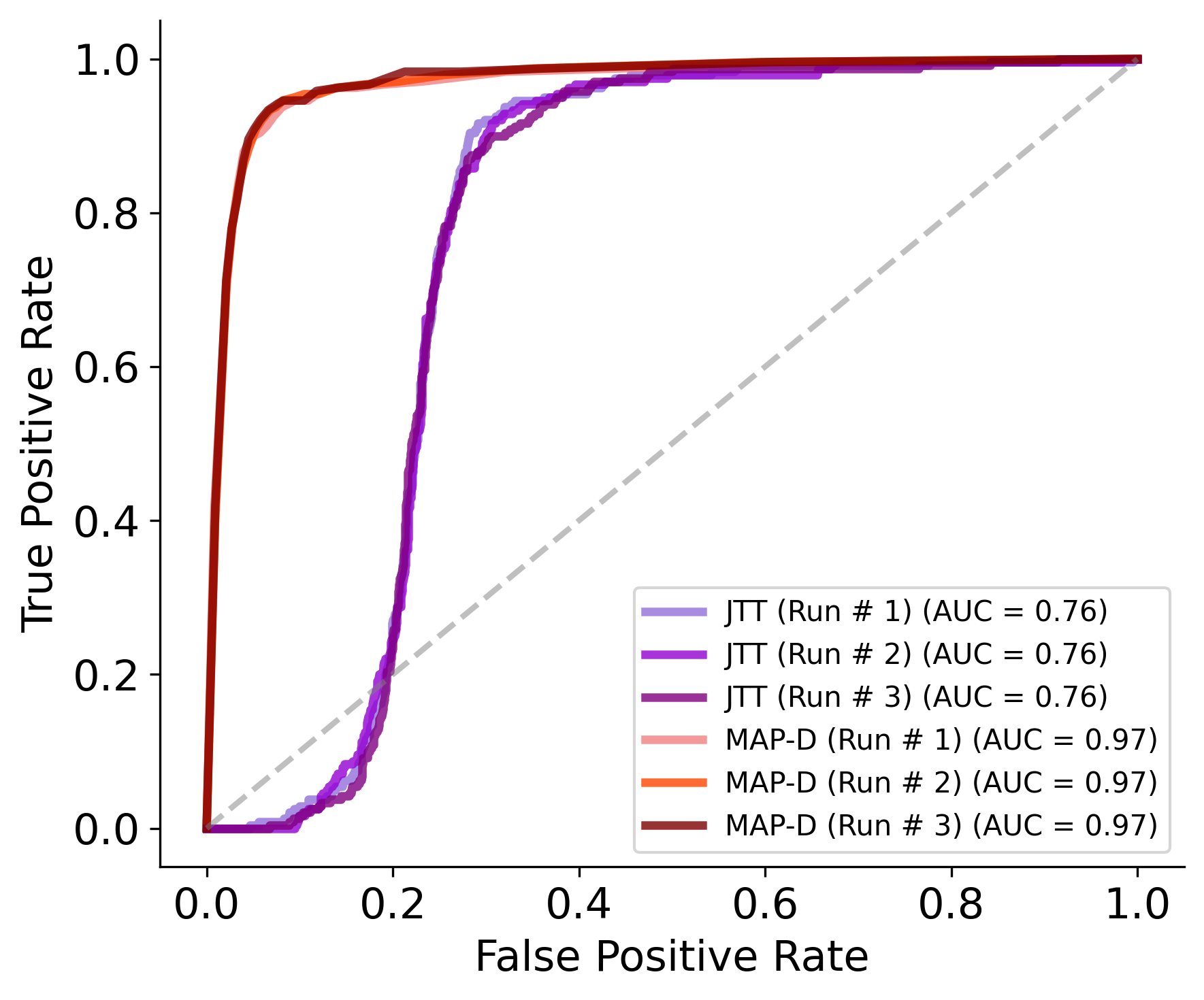}
    }
    \hspace{8mm}
    \subfloat[CelebA]{
        \includegraphics[width=0.4\linewidth]{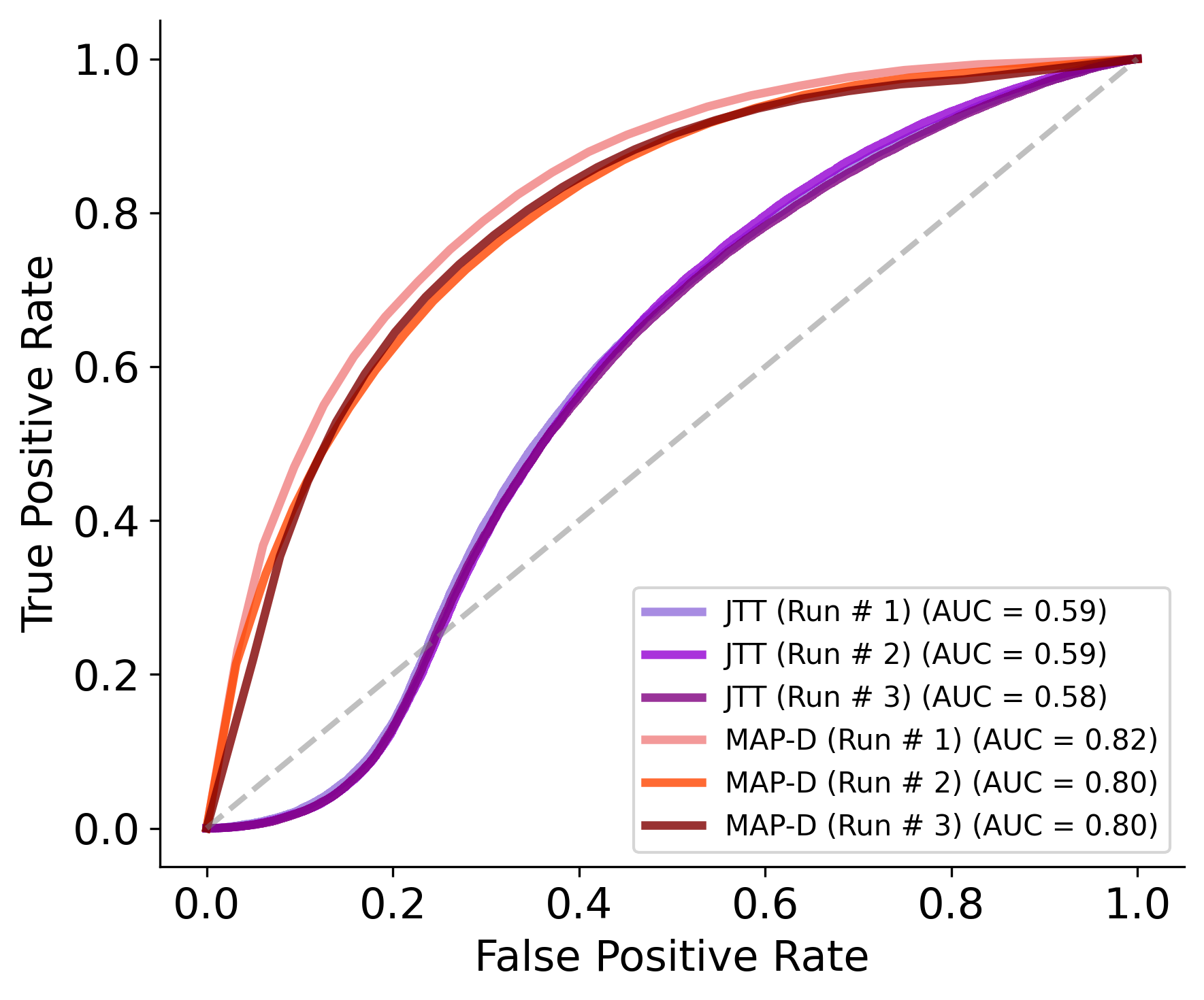}
    }
    \caption{Demonstration of the effectiveness of \textit{MAP-D} in detecting minority-group samples on two famous minority-group datasets with spurious correlations, compared to the detection performance of JTT~\cite{liu2021just} which relies on early-stopping. \textit{MAP-D} achieves better or similar performance, without needing costly hyperparameter tuning or retraining.}
    \label{fig:minority_group}
\end{figure*}

Minimizing average-case error often hurts performance on minority sub-groups that might be present in a dataset~\cite{sagawa2019distributionally,sagawa2020investigation,liu2021just}.
For instance, models might learn to rely on spurious features that are only predictive for majority groups.
Identifying minority-group samples can help detect and correct such issues, improving model fairness.

Previous works identify minority examples as those that are not already fit after some number of training epochs, and retrain from scratch with those examples upweighted \citep{liu2021just,zhang2022correctncontrast}.
The number of epochs is treated as a hyperparameter; tuning it requires running the training process twice (without and then with upweighting) and evaluating on held-out known-to-be-minority examples.
Instead of relying on the inductive bias that minority examples will be harder to fit, we apply \textit{MAP-D} to find examples that match minority examples' training dynamics, and find this is much more effective method of identifying minority examples, see Figure
~\ref{fig:minority_group}. 
This avoids the costly hyperparameter tuning required by previous methods.
Instead of just using 250 examples per probe category, we use the complete validation set in this case to be consistent with prior work~\cite{liu2021just}.
Furthermore, these examples are not included as part of the training set in order to match the statistics of examples at test time.

\section{Related Work}
\label{sec:related}

Many research directions focus on properties of data and leveraging them in turn to improve the training process. We categorize and discuss each of these below.

\paragraph{Difficulty of examples} 
\citet{2017koh} proposes influence functions to identify training points most influential on a given prediction. Work by \citet{arpit2017closer,li2020gradient,feldman2019does,feldman2020neural} develop methods that measure the degree of memorization required of individual examples.
While \citet{2020Jiang} proposes a consistency score to rank each example by alignment with the training instances, \citet{2019carlini} considers several different measures to isolate prototypes that could conceivably be extended to rank the entire dataset. \citet{agarwal2021estimating} leverage variance of gradients across training to rank examples by learning difficulty. Further, \citet{2019arXiv191105248H} classify examples as challenging according to sensitivity to varying model capacity. In contrast to all these approaches that attempt to rank an example along one axis, \textit{MAP-D} is able to discern between different sources of uncertainty without any significant computational cost by directly leveraging the training dynamics of the model.

\paragraph{Coreset selection techniques} The aim of these methods is to find prototypical examples that represent a larger corpus of data~\citep{Zhang1992,2012Bien,2015kim,NIPS2016_6300}, which can be used to speed up training~\citep{sener2018active,Shim2021CoresetSF,huggins2017coresets} or aid in interpretability of the model predictions \citep{yoon2019rllim}.
\textit{MAP-D} provides a computationally feasible alternate to identify and surface these coresets.

\paragraph{Noisy examples} A special case of example difficulty is noisy labels, and correcting for their presence.
\citet{arazo2019unsupervised} use parameterized mixture models with two modes (for clean and noisy) fit to sample loss statistics, which they then use to relabel samples determined to be noisy.
\citet{li2020dividemix} similarly uses mixture models to identify mislabelled samples, but actions on them by discarding the labels entirely and using these samples for auxiliary self-supervised training.
These methods are unified by the goal of identifying examples that the model finds challenging, but unlike \textit{MAP-D}, do not distinguish between the sources of this difficulty.

\paragraph{Leveraging training signal} There are several prior techniques that also leverage network training dynamics over distinct phases of learning~\citep{Achille2017CriticalLP,2020Jiang,Mangalam2019DoDN,2020fartash,agarwal2021estimating}. Notably, \citet{pleiss2020identifying} use loss dynamics of samples over the course of training, but calculate an Area-Under-Margin metric and show it can distinguish correct but difficult samples from mislabelled samples. In contrast, \textit{MAP-D} is capable of inferring multiple data properties.
Swayamdipta et al. (2020)~\cite{swayamdipta2020dataset} computed the mean and variance of the model's confidence for the target label throughout training to identify interesting examples in the context of natural language processing. However, their method is limited in terms of identifying only easy, hard, or confusing examples. Our work builds upon this direction and can be extended to arbitrary sources of uncertainty based on defined probe suites leveraging loss trajectories.

\paragraph{Adaptive training} 
Adaptive training leverages training dynamics of the network to identify examples that are worth learning.
Loss-based prioritization \citep{jiang2019,Katharopoulos2018} upweight high loss examples, assuming these examples are challenging yet learnable.
These methods have been shown to quickly degrade in performance in the presence of even small amounts of noise since upweighting noisy samples hurts generalization \citep{hu2021does,paul2021deep}. \Citet{dsouza2021} motivate using targeted data augmentation to distinguish between different sources of uncertainty, and adapting training based upon differences in rates of learning. On the other hand, several methods prioritize learning on examples with a low loss assuming that they are more meaningful to learn. Recent work has also attempted to discern between points that are learnable (not noisy), worth learning (in distribution), and not yet learned (not redundant)~\citep{mindermann2022prioritized}.
\textit{MAP-D} can also be leveraged for adaptive training by defining the different sources of uncertainties of interest.

\paragraph{Minority group samples} 
The recent interest has been particularly towards finding and dealing with minority group samples to promote model fairness~\cite{sagawa2019distributionally,sagawa2020investigation,liu2021just,zhang2022correctncontrast,nam2022spreadspurious}.
The dominant approach to deal with this problem without assuming access to group labels is to either pseudo-label the dataset using a classifier~\cite{nam2022spreadspurious} or to train a model with early-stopping via a small validation set to surface minority group samples~\cite{liu2021just,zhang2022correctncontrast}.
However, this setting only works for the contrived datasets where the model can classify the group based on the background.
\textit{MAP-D} leverages the population statistics rather than exploiting the curation process of the dataset to naturally surface minority group samples, which is scalable and applicable in the real-world.

\section{Conclusion} \label{sec:conclusion}

We introduce the problem of \textit{Metadata Archeology} as the task of surfacing and inferring metadata of different examples in a dataset, noting that the relational qualities of metadata are of special interest (as compared to ordinary data features) for auditing, fairness, and many other applications. 
Metadata archaeology provides a unified framework for addressing multiple such data quality issues in large-scale datasets. We also propose a simple approach to this problem, \textit{Metadata Archaeology via Probe Dynamics (\textit{MAP-D})}, based on the assumption that examples with similar learning dynamics present the same metadata. We show that \textit{MAP-D} is successful in identifying appropriate metadata features for data examples, even with no human labelling, making it a competitive approach for a variety of downstream tasks and datasets and a useful tool for auditing large scale datasets.

\paragraph{Limitations} This work is focused on a computer vision setting; we consider an important direction of future work to be extending this to other domains. \textit{MAP-D} surfaces examples from the model based on the loss trajectories. This is based on a strong assumption that these loss trajectories are separable. It is possible that the learning curve for two set of probe categories exhibit similar behavior, limiting the model's capacity in telling them apart. In this case, the learning curve is no longer a valid discriminator between probes. However, for good constructions of probe categories relying on global population statistics, we consider \textit{MAP-D} to be a competitive and data-efficient method. 

\newpage
\bibliography{references.bib}

\newpage

\appendix

\section*{Appendix}

\section{Experimental Details}\label{sec:expdetails}

In all experiments, we use variants of the ResNet architecture and leverage standard image classification datasets -- CIFAR-10/100 and ImageNet. We train with SGD using standard hyperparameter settings: learning rate 0.1, momentum 0.9, weight-decay 0.0005, and a cosine learning rate decay. We achieve top-1 accuracies of 93.68\% on CIFAR-10, 72.80\% on CIFAR-100, and 73.94\% on ImageNet.

\paragraph{CIFAR-10/100} To account for the smaller image size in this dataset, we follow standard practice and modify the models input layer to have stride 1 and filter size 3. We use a batch-size of 128 and train for 150 epochs. We use random horizontal flips and take a random crop of size $32 \times 32$ after padding the image using reflection padding with a padding size of 4~\cite{he2016deep}.
For label noise correction experiments, we follow the experimental protocol of~\cite{arazo2019unsupervised} with ResNet-18 where we train the model for 300 epochs with SGD and an initial learning rate of 0.1 decayed by a factor of 0.1 at the 100\textsuperscript{th} and 250\textsuperscript{th} epoch. A weight decay of 0.0001 is also applied.

\paragraph{ImageNet} We use a batch-size of 256 and train for 100 epochs. We apply center crop augmentation for testing as per the common practice (i.e. resize image to $256 \times 256$ and take the center crop of size $224 \times 224$) ~\cite{he2016deep,nvidia_resnet50_v1.5}.

\paragraph{Waterbirds / CelebA} We use the same model architecture and hyperparameters as~\cite{liu2021just} in order to enable a fair and direct comparison. All experiments are based on the default ResNet-50 architecture. The Waterbirds models are trained for 300 epochs using SGD with an initial learning rate of 0.00001, and a high weight decay of 1.0. The model was early-stopped after the 60\textsuperscript{th} epoch for JTT~\cite{liu2021just}. The CelebA models are trained for 50 epochs using SGD with an initial learning rate of 0.00001, and a high weight decay of 0.1. The model was early-stopped after the first epoch for JTT~\cite{liu2021just}.

\paragraph{Clothing1M} We use the online batch selection protocol from~\cite{mindermann2022prioritized} where 32 examples are chosen from a large batch of 320 examples for training at each step.
Following~\cite{mindermann2022prioritized}, we use AdamW optimizer with default hyperparameters as in PyTorch~\cite{paszke2019pytorch} and ImageNet pretrained ResNet-50.
No learning rate decay is applied in this case.

\section{Probe suites for CIFAR-100}

\begin{figure*}[!t]
    \centering
    \subfloat[Typical]{
        \includegraphics[width=0.32\linewidth]{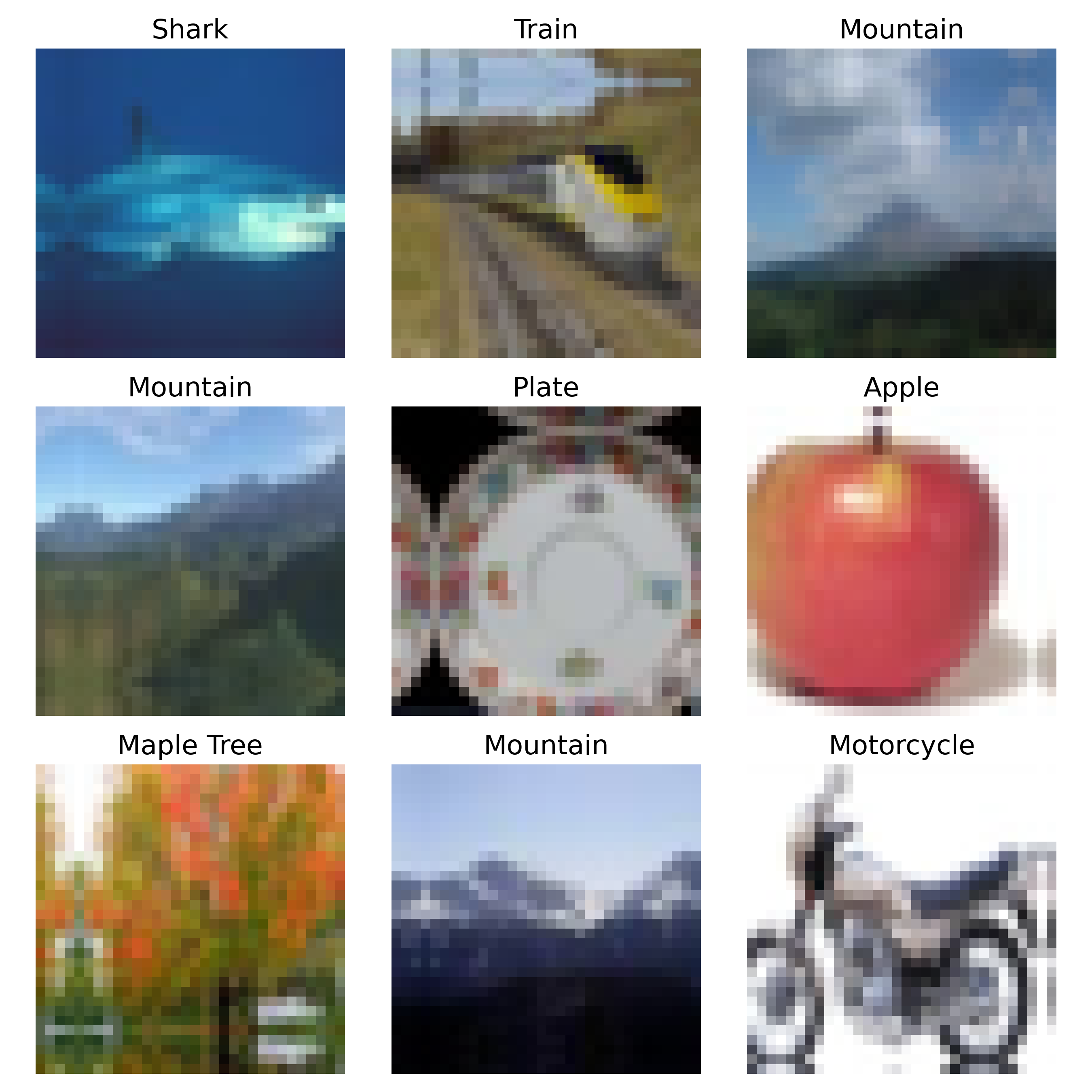}
        \label{fig:probes_typical_c100}
    }
    \subfloat[Atypical]{
        \includegraphics[width=0.32\linewidth]{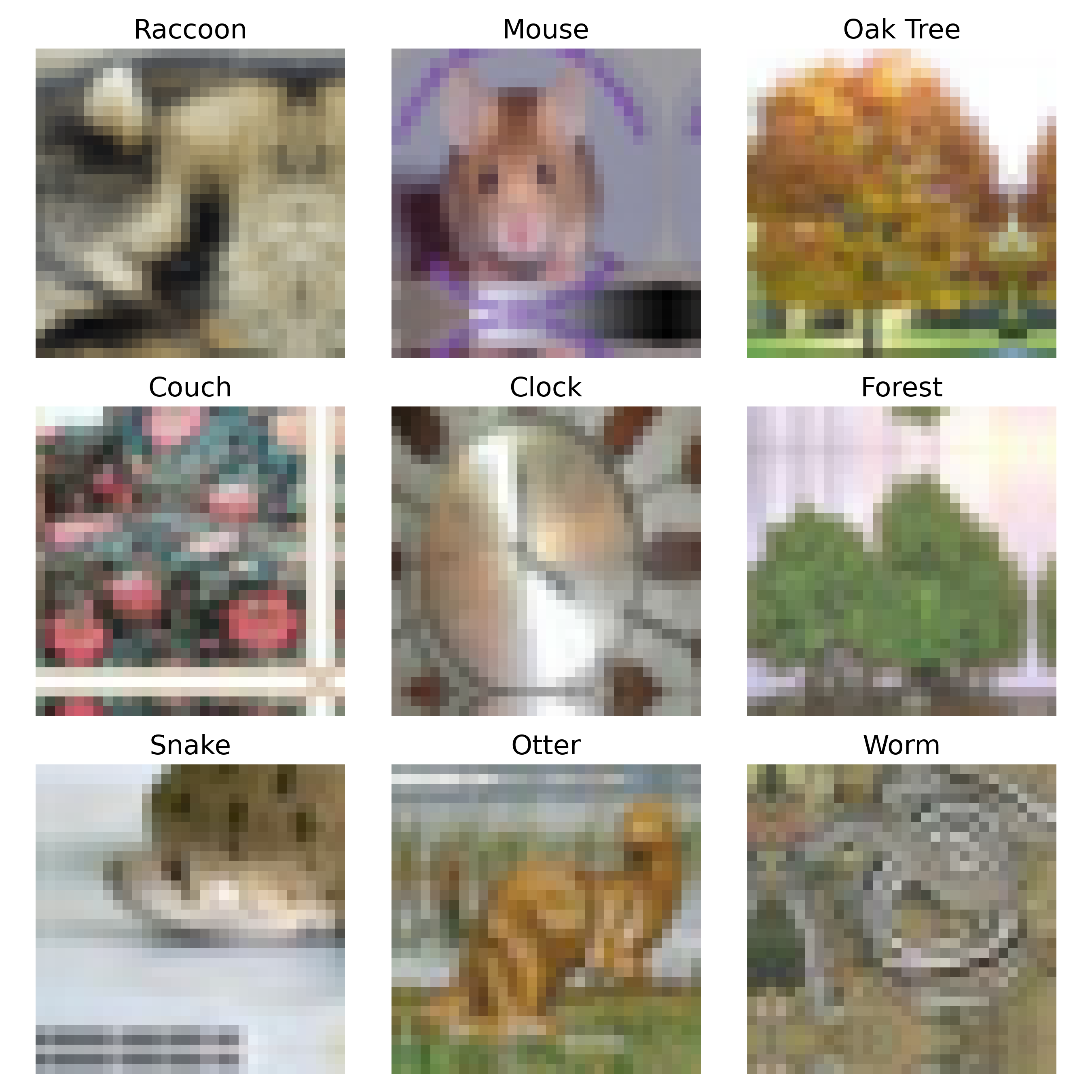}
        \label{fig:probes_atypical_c100}
    }
    \subfloat[Corrupted]{
        \includegraphics[width=0.32\linewidth]{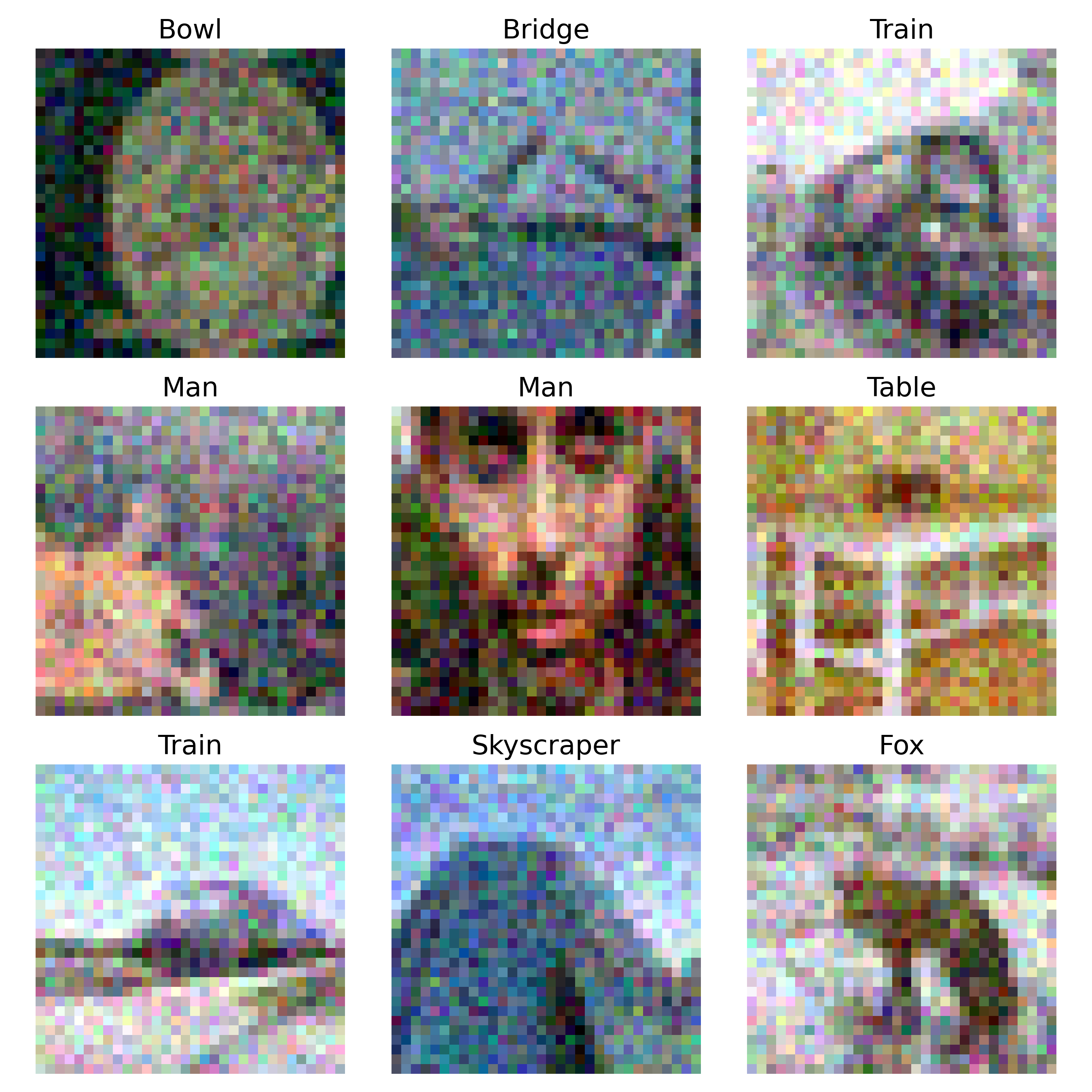}
        \label{fig:probes_corrupted_c100}
    }
    
    \subfloat[Out-of-Distribution]{
        \includegraphics[width=0.32\linewidth]{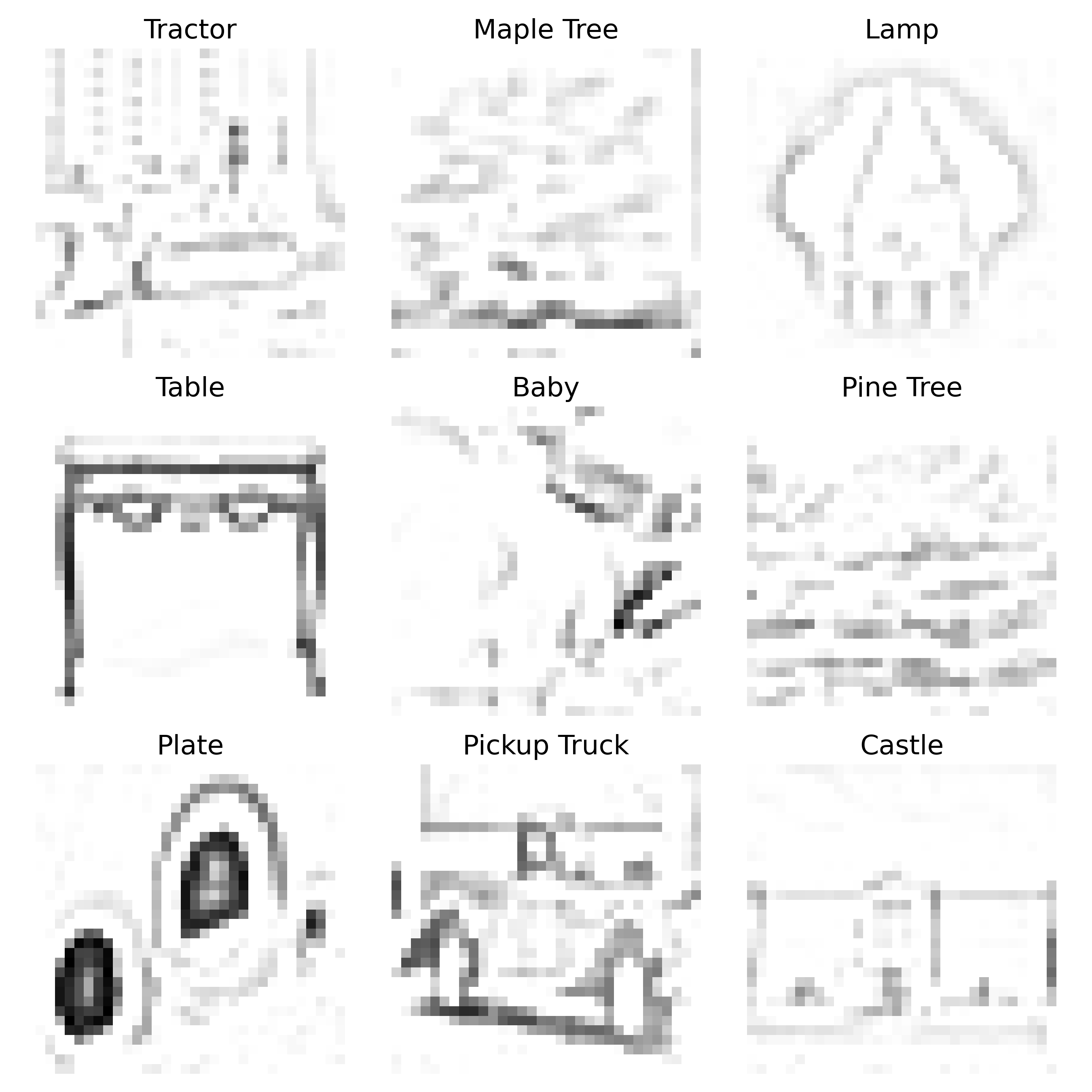}
        \label{fig:probes_ood_c100}
    }
    \subfloat[Random Outputs]{
        \includegraphics[width=0.32\linewidth]{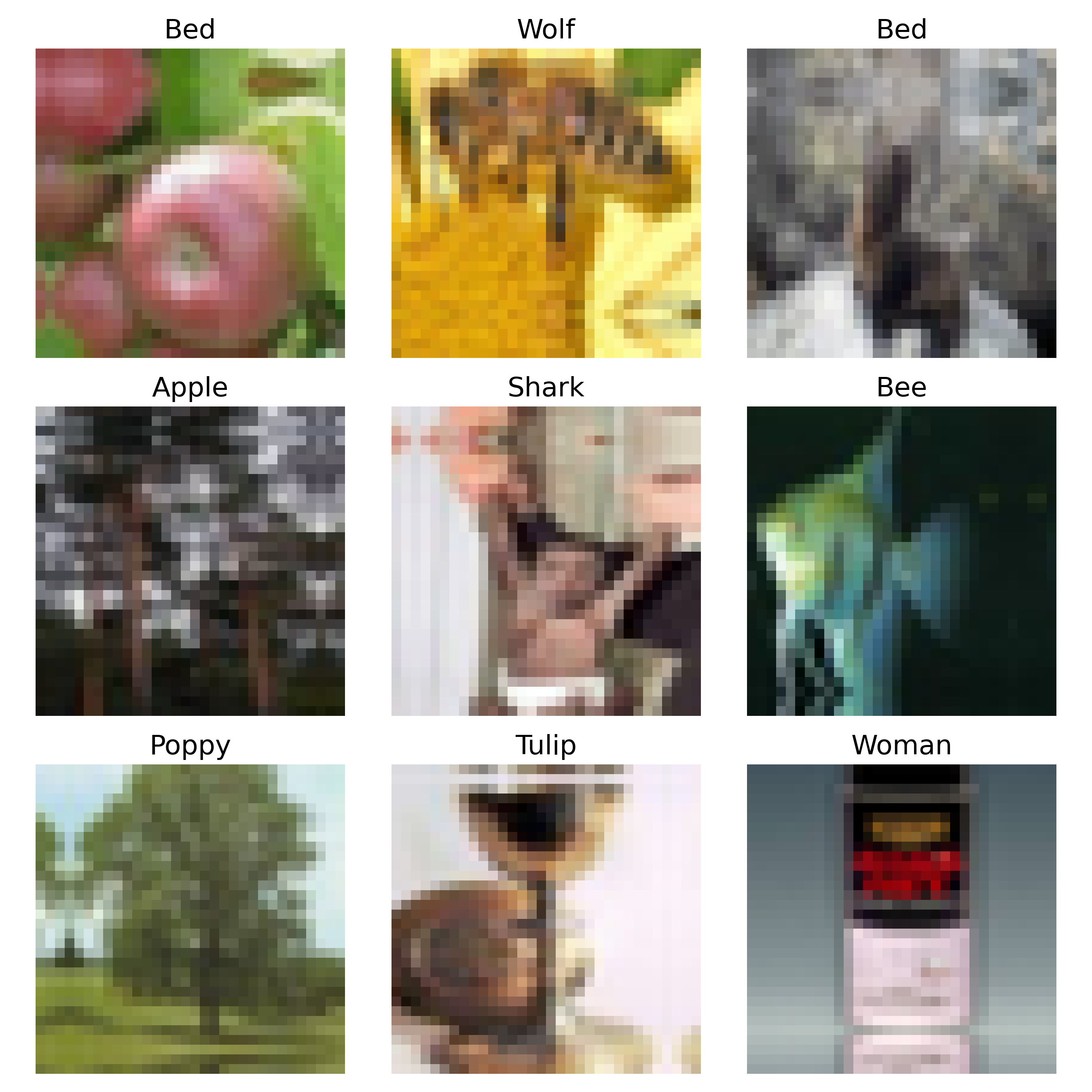}
        \label{fig:probes_mislabeled_c100}
    }
    \subfloat[Random Inputs \& Outputs]{
        \includegraphics[width=0.32\linewidth]{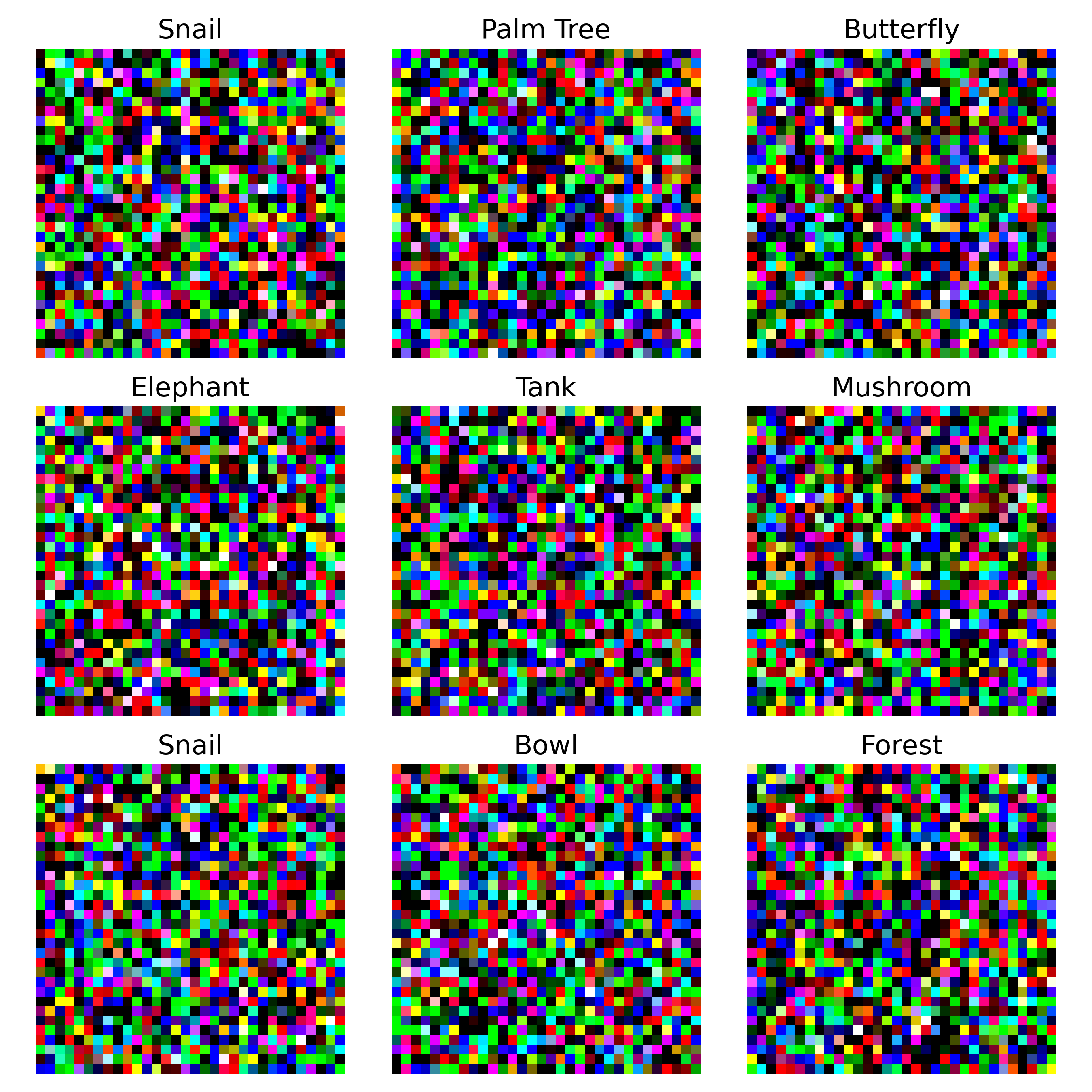}
        \label{fig:probes_noisy_c100}
    }
    
    \caption{An example of the different kind of probes that can be defined on the CIFAR-100 dataset.}
    \label{fig:probe_suites_c100}
\end{figure*}

We present examples from the curated probe suites on CIFAR-100 in Fig.~\ref{fig:probe_suites_c100}.

\begin{figure*}[!t]
    \centering
    \includegraphics[width=0.48\linewidth]{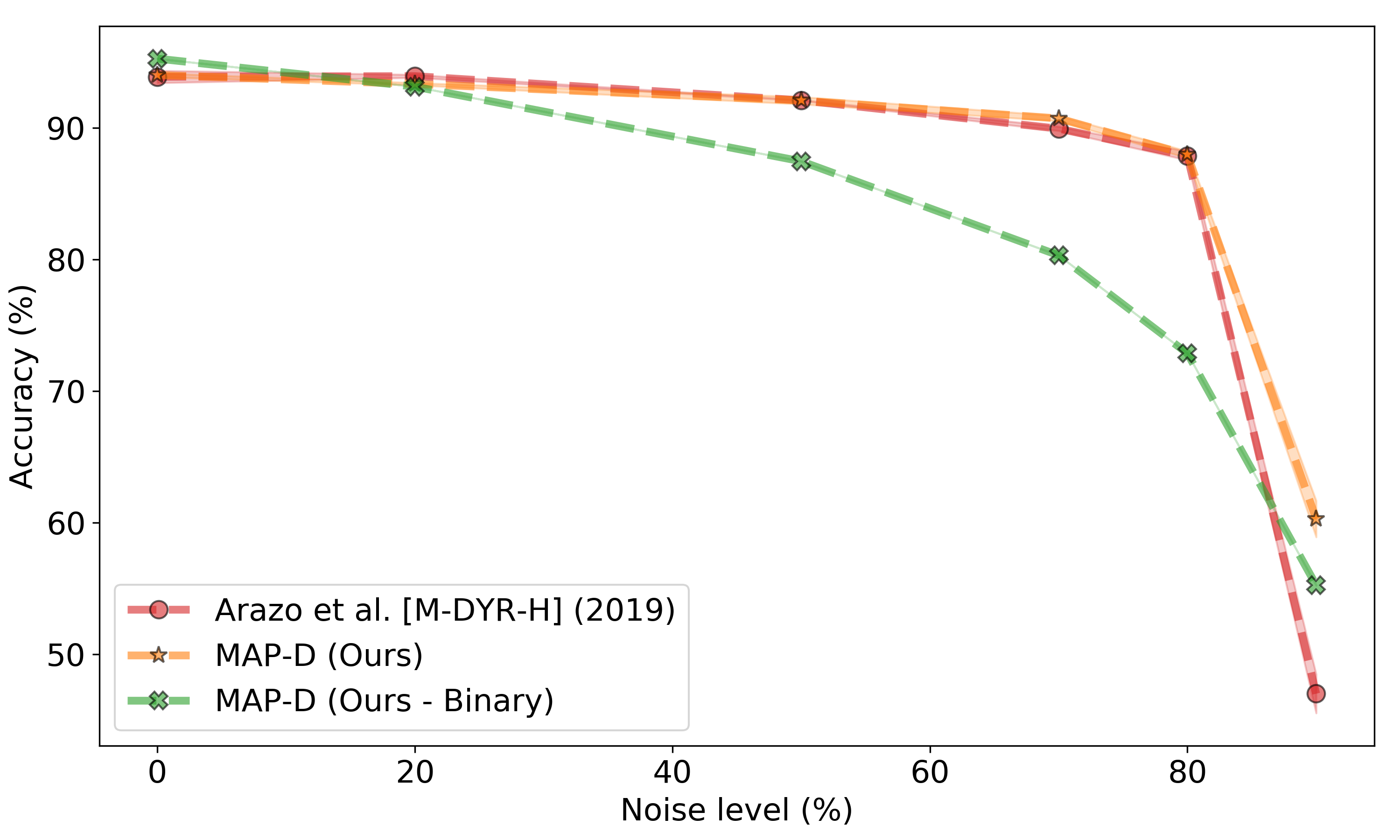}
    \caption{Ablation for label correction on CIFAR-10, where we use a binary prediction instead of probability estimates returned by \textit{MAP-D}. This highlights the utility and effectiveness of the uncertainty estimates computed by \textit{MAP-D}.}
    \label{fig:ablation_label_correction_c10}
\end{figure*}

\section{Binary vs. Probabilistic Outputs in Label Correction} \label{sub_sec:prob_outputs_label_corr}
Arazo et al. (2019)~\cite{arazo2019unsupervised} used a convex combination of the labels weighted by the actual probability returned by their BMM model.
As \textit{MAP-D} returns probability estimates, this enabled leveraging label correction framework in the same way.
However, the utility of the uncertainty estimates is not immediately apparent.
Therefore, in order to gauge the utility of these uncertainty estimates, we used binary predictions (argmax) instead of the actual probabilities returned by \textit{MAP-D}.
The results are visualized in Fig.~\ref{fig:ablation_label_correction_c10}.
It is clear from the figure that the model struggles significantly in coping with noise when being restricted to binary predictions, indicating that the uncertainty estimates provided by \textit{MAP-D} enables the model to learn the correct label.

\end{document}